\documentclass[journal]{IEEEtran}
\usepackage{graphicx}
\usepackage{amsmath,amssymb} 
\usepackage{color}
\usepackage{array}
\usepackage{epstopdf}
\usepackage{algorithm}
\usepackage{algorithmic}
\usepackage{bm}
\usepackage{multirow}
\usepackage{multirow,array}
\usepackage[pagebackref=true,breaklinks=true,letterpaper=true,colorlinks,bookmarks=false]{hyperref}
\ifCLASSINFOpdf
\else
\fi

\begin{document}
%
\title{Hybrid CNN and Dictionary-Based Models for Scene Recognition and Domain Adaptation}
%
%
%

\author{Guo-Sen Xie,
	    Xu-Yao Zhang,
        Shuicheng Yan,~\IEEEmembership{Senior Member,~IEEE,}
        Cheng-Lin Liu,~\IEEEmembership{~Fellow,~IEEE}%
\thanks{G.-S. Xie, X.-Y. Zhang and C.-L. Liu are with the National Laboratory of Pattern Recognition, Institute of Automation, Chinese Academy of Sciences, No. 95 Zhongguancun East Road, Beijing 100190, P.R. China. C.-L. Liu is also with the Research Center for Brain-Inspired Intelligence, Institute of Automation, and the CAS Center for Excellence in Brain Science and Intelligence Technology, Chinese Academy of Sciences. E-mail: \{guosen.xie, xyz, liucl\}@nlpr.ia.ac.cn. }

\thanks{S.~Yan is with the Department of Electrical and Computer Engineering, National University of Singapore, 117583, Singapore. Email: eleyans@nus.edu.sg.}}

%
%

\markboth{Transaction on circults and systems for video technology}%
{Shell \MakeLowercase{\textit{et al.}}: Bare Demo of IEEEtran.cls for Journals}

%



\maketitle

\begin{abstract}
Convolutional neural network (CNN) has achieved state-of-the-art performance in many different visual tasks. Learned from a large-scale training dataset, CNN features are much more discriminative and accurate than the hand-crafted features. Moreover, CNN features are also transferable among different domains. On the other hand, traditional dictionary-based features (such as BoW and SPM) contain much more local discriminative and structural information, which is implicitly embedded in the images. To further improve the performance, in this paper, we propose to combine CNN with dictionary-based models for scene recognition and visual domain adaptation. Specifically, based on the well-tuned CNN models~(e.g., AlexNet and VGG Net), two dictionary-based representations are further constructed, namely mid-level local representation (MLR) and convolutional Fisher vector representation (CFV). In MLR, an efficient two-stage clustering method, i.e., weighted spatial and feature space spectral clustering on the parts of a single image followed by clustering all representative parts of all images, is used to generate a class-mixture or a class-specific part dictionary. After that, the part dictionary is used to operate with the multi-scale image inputs for generating mid-level representation. In CFV, a multi-scale and scale-proportional GMM training strategy is utilized to generate Fisher vectors based on the last convolutional layer of CNN. By integrating the complementary information of MLR, CFV and the CNN features of the fully connected layer, the state-of-the-art performance can be achieved on scene recognition and domain adaptation problems. {\color{black}An interested finding is that our proposed hybrid representation~(from VGG net trained on ImageNet) is also complementary with GoogLeNet and/or VGG-11~(trained on Place205) greatly.} 
\end{abstract}


%
\IEEEpeerreviewmaketitle

\section{Introduction}
%
%
%
%
\IEEEPARstart{W}{ith} the development of deep learning, convolutional neural network~(CNN)~\cite{lecun1998gradient},~\cite{krizhevsky2012imagenet} has been successfully applied to various fields, such as object recognition~\cite{krizhevsky2012imagenet},~\cite{SimonyanZ14a},~\cite{szegedy2014going},~\cite{BatchNormalize},~\cite{MSRA}, image detection~\cite{girshick2014rich},~\cite{sermanet2013pedestrian},~\cite{ouyang2014deepid}, image segmentation~\cite{kang2014fully},~\cite{LongSD14},~\cite{PinheiroC14},~\cite{ChenPKMY14}, image retrieval~\cite{Zhao14}, and so on. State-of-the-art performance achieved by CNN in these fields identify the powerful feature representation ability of CNN for different visual tasks.

The traditional Bag of Visual Words (BoW) models~\cite{jurie2005creating},~\cite{yang2009linear},~\cite{wang2010locality},~\cite{perronnin2007fisher},~\cite{perronnin2010improving},~\cite{zhou2010image}, which had been quite popular before 2012 in the research community of object recognition, is now being gradually neglected since the CNN model gained the champion in the large-scale competition of ImageNet classification~\cite{krizhevsky2012imagenet} in 2012~(ILSVRC-12). Due to the strong representation ability of CNN trained on a large dataset, e.g., ImageNet, some works~\cite{razavian2014cnn},~\cite{donahue2013decaf} advocate directly using CNN features for classification, and have gained much better performance than traditional methods. There also exist some works~\cite{zuo2014learning},~\cite{gong2014multi} that have tried to combine traditional models with the CNN model to get better results. Traditional features and CNN features can complement each other, and better performance can be obtained by combining them.

The images from the ImageNet database for training CNN are mostly object-oriented, thus it seems that the trained CNN model is more suitable for object recognition than for other tasks, e.g., scene recognition~\cite{quattoni2009recognizing}. Zhou et al.~\cite{zhou2014learning}  collected a large-scale place database to train the CNN (PlaceNet), with the same architecture as~\cite{donahue2013decaf}, and found that based on the features of PlaceNet, better performance can be obtained than the original CNN features~\cite{donahue2013decaf}. Recently, Mircea et al.~\cite{cimpoi2014deep} found that stronger CNN architectures can approach and outperform PlaceNet even if trained on ImageNet data. {\color{black}{Yosinski et al.~\cite{NIPS2014_5347} verified that CNN features in the lower layer are more ``general" while the features of higher layers are more ``specific".}} Therefore, how to explore the underlined representation and discriminative ability of CNN, no matter what kind of database it is trained on, is still an interesting problem.
\begin{figure}[t!]
	\centering
	\includegraphics[width=0.5\textwidth]{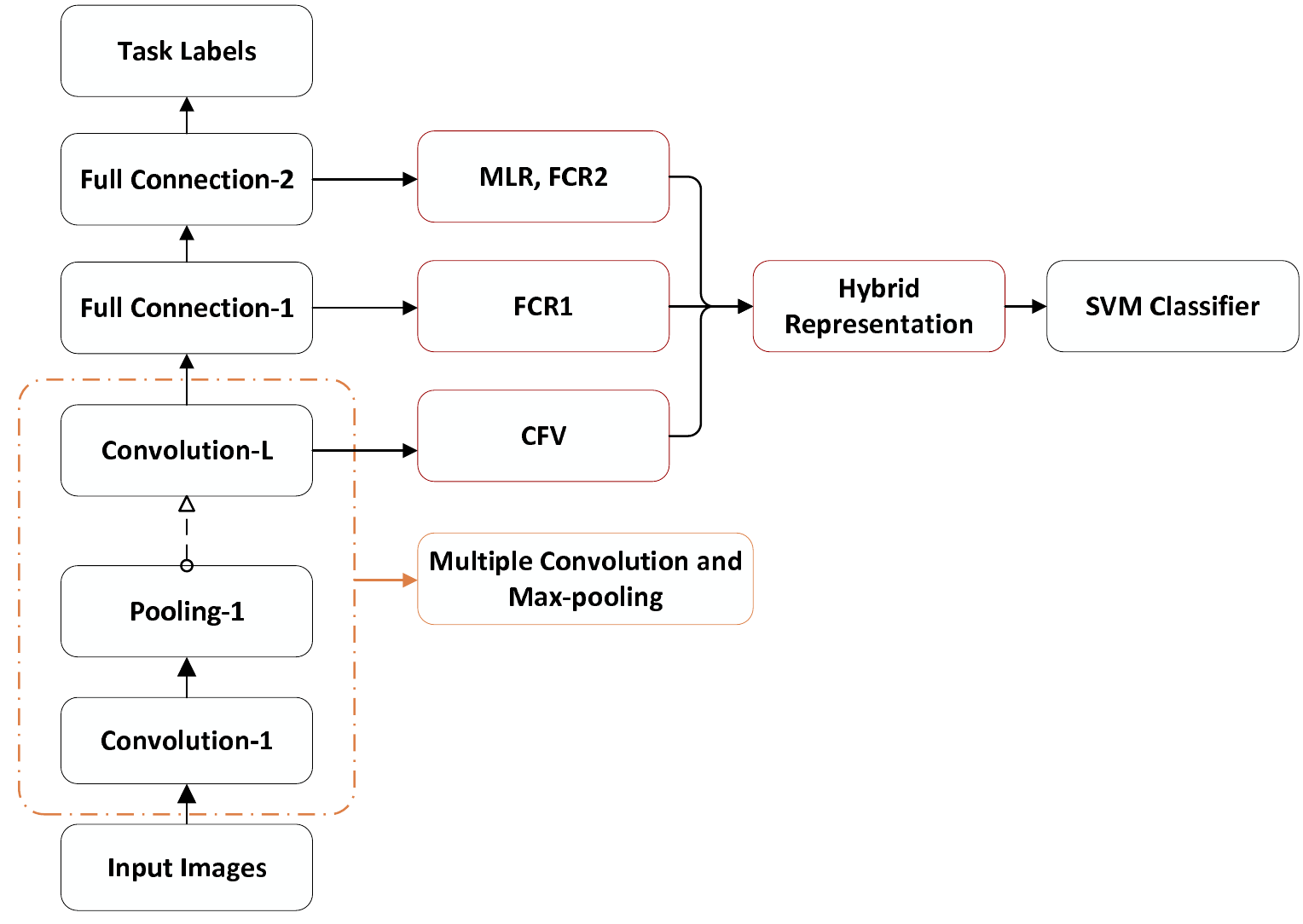}
	\caption{The proposed hybrid representation based on the trained CNN. CFV is calculated based on the last convolutional layer~(Convolution-L); FCR1 indicates the global representation based on the first fully connected layer~(Full Connection-1); similarly, FCR2 is calculated based on the second fully connected layer~(Full Connection-2); and our proposed MLR is based on the second fully connected layer too. Operations in the dashed box are multiple convolution, max-pooling and other normalization, determined by the used CNN architecture.}
	\label{fig1}
\end{figure}
In this paper, we focus on scene recognition~\cite{quattoni2009recognizing} and visual domain adaptation~\cite{saenko2010adapting}. Unlike in generic object categorization, scene images have many discriminative part regions, which is beneficial for distinguishing the categories. During the past few years, many works~\cite{juneja2013blocks},~\cite{singh2012unsupervised},~\cite{doersch2013mid},~\cite{lin2014learning},~\cite{doersch2012makes},~\cite{pandey2011scene} have been devoted to discover discriminative parts for scene recognition. The above methods first train a detector or classifier and then detect the part regions, which are not scalable for large databases. To discover discriminative parts more efficiently, there have also been some methods that are based on bottom cues to discover discriminative proposals, e.g.,~\cite{uijlings2013selective},~\cite{cheng2014bing},~\cite{arbelaez2014multiscale},~\cite{zitnick2014edge}. Visual domain adaptation~(DA) can be classified into two types,  unsupervised and semi-supervised adaptation, according to whether the target domain samples are used or not for training the classifier. Pioneer works of DA include~\cite{saenko2010adapting},~\cite{gopalan2011domain},~\cite{gong2012geodesic},~\cite{gong2013connecting},~\cite{fernando2013unsupervised},~\cite{chopra2013dlid}, which are based on learning with regularization on the manifold. Recent works on DA~\cite{tzeng2014deep},~\cite{Menglong2015deep},~\cite{ghifary2014domain},~\cite{borgwardt2006integrating} have also adopted the strategy of adding maximum mean discrepancy~(MMD) constraint on the fully connected layer of neural networks during the training process.

{\color{black}For both} scene recognition and domain adaptation, very few works have  utilized local discriminative information implicitly contained in the images in the context of CNN. Especially for domain adaptation, no one knows whether the local part information can improve the domain transfer performance or not. Herein, we propose to utilize the local part information to improve the discriminative ability of CNN features. Specifically, we propose to combine CNN features with two dictionary-based models. {\color{black}{The first one is the the mid-level local discriminative representation~(MLR). MLR aims to discover local discriminative information contained in the images, which are beneficial for afterward classification. To construct MLR, we utilize selective search~\cite{uijlings2013selective} to generate initial parts for each image, followed by our proposed two-stage clustering to filter out redundancy parts and generate part dictionary, finally the locality-constrained linear coding~(LLC)~\cite{wang2010locality} and spatial pyramid matching (SPM)~\cite{lazebnik2006beyond} are used for generating the MLR.~(See Fig.~\ref{fig2} for illustration.)}} 
On the other hand, Fisher vectors of the last convolutional layer of CNN~(CFV) are generated to further boost the performance. {\color{black}{As we know, Fisher vector~\cite{perronnin2007fisher} consists of first order and second order differences between the descriptors and the GMM centers, which gives CFV the same representation ability to well distinguish different categories.}} As for GMM training before Fisher vector coding, muti-scale and scale-proportional descriptors are sampled. As for Fisher vector coding, we use the same strategy as~\cite{yoo2014fisher}, which is denoted as Multi-scale Pyramid Pooling. Another commonly used feature of CNN is the fully connected layer representation~(FCR), which contains the global information of input images. By combining these several representations, i.e., mid-level local representation~(MLR), convolutional Fisher vector representation~(CFV), and the global representions of the last two fully connected layers of CNN~(FCR) together, we can obtain our hybrid representation~(See Fig.~\ref{fig1} for the whole flowchart). {\color{black}{The advantages of our hybrid representation lie in~1) having more discriminative ability for classification, and 2) being complementary to each other.}} Experimental results on both scene recognition and domain adaptation validate the strong domain transfer ability from other large database of our hybrid representation. Moreover, we also find that MLR is both complementary with CFV and FCR.

The remainder of this paper is organized as follows. In Section~II, we give some related works. In Section~III, we illustrate the proposed pipeline elaborately. Section~IV  shows that the hybrid representation can achieve the best results in scene recognition and visual domain adaptation. In Section~V, we conclude this paper and discuss the future work.

\section{Related Work}
In this section, we introduce the related works to our hybrid representation. Gong et al.~\cite{gong2014multi} proposed to calculate the VLAD~\cite{jegou2010aggregating} representation based on the FCR of the trained CNN, of which the performance is better than the original FCR and other traditional methods. Let $X=[x_{1},x_{2},\cdots,x_{N}]\in \mathbb{R}^{d\times N}$  be the single-scale activates of the image $I$ based on the FCR of the trained CNN, and $C=[c_{1},c_{2},\cdots,c_{M}]\in \mathbb{R}^{d\times M}$ be the codebook (dictionary) learned by the K-means algorithm based on all the training images. Then, the VLAD coding of the image $I$ can be represented as
\begin{equation}
v = [\sum\limits_{NN(x_{i})=c_{1}} x_{i}-c_{1},\cdots, \sum\limits_{NN(x_{i})=c_{M}} x_{i}-c_{M}],
\label{Eq1}
\end{equation}
where $NN(x_{i})=c_{j}$ denotes the set of $x_{i}$ whose nearest neighbor is $c_{j}$. To improve the performance, Gong et al.~\cite{gong2014multi} calculated multi-scale representations of Eq.~\ref{Eq1} and concatenated all the scales. Motivated by~\cite{gong2014multi}, the CFV~\cite{perronnin2007fisher} can be further used to improve the performance of image representation~\cite{cimpoi2014deep},~\cite{yoo2014fisher}.

Suppose we are given the multi-scale activates \{${X^{(s)}|X^{(s)}\in \mathbb{R}^{d\times N^{(s)}}}$\}, $X^{(s)}=[x_{1}^{(s)},x_{2}^{(s)},\cdots,x_{N^{(s)}}^{(s)}]$,~$s = 1,2,\cdots,S$ from all the training images. Let $u_{\lambda}=\sum_{t=1}^{M}\omega_{i}u_{i}(x)$  denote a Gaussian Mixture Model~(GMM), where $\lambda=\{\lambda_{i}=\{\omega_{i},\mu_{i},\sigma_{i}\},i=1,2,\cdots,M\}$ represents the parameters of the GMM, and $\lambda$ can be optimized by the Maximum Likelihood (ML) estimation based on \{$X^{(s)},s=1,2,\cdots,S$\}. Denote the multi-scale activates of the  image $I$ as $\chi=\{\chi_{s}=[x_{1}^{(s)},x_{2}^{(s)},\cdots,x_{I^{(s)}}^{(s)}],s=1,2,\cdots,S\}$. The gradients of the  $u_{\lambda}(\chi)$ w.r.t. the $i$-th Gaussian can be represented as follows
\begin{equation}
g_{i} = \frac{1}{|\chi|}\sum\limits_{s=1}^{S}\sum\limits_{n=1}^{|\chi_{s}|}\nabla_{\lambda_{i}}\log u_{\lambda}(x_{n}^{(s)}).
\label{Eq2}
\end{equation}
The Fisher vector of the image $I$ is obtained by concatenating  all the gradients w.r.t. those $M$ Gaussians.
In~\cite{yoo2014fisher},  while calculating gradients w.r.t. each Gaussian, Yoo et al. adopted Multi-scale Pyramid Pooling, where first the GMM parameters are generated based on all the descriptors from different scaled training images, and then scale-specific normalization and max-pooling are implemented.

Recently, Liu et al.~\cite{Liu2014ACCV} also proposed to learn a part dictionary based on the LC-KSVD method~\cite{jiang2011learning} directly. In this work, each part element is given a label which is the same as that of the image where it is located on. LC-KSVD is a popular dictionary learning method, but it is very time-consuming and therefore cannot address large-scale problems, e.g., the part dictionary learning of the SUN database~\cite{xiao2010sun}. Another drawback is that many local parts usually have no explicit semantic~(label) information at all, so unsupervised part dictionary learning may be more reasonable than the supervised counterpart.

\section{The Proposed Hybrid Method}
In this section, we illustrate the whole pipeline of the construction of MLR, CFV and FCR, followed by the details of each part.
\subsection{The Whole System}
To explore the representation ability of deep CNN features, we propose to combine the three kinds of representations based on CNN features, i.e., our proposed MLR, CFV, and FCR. We illustrate the whole pipeline in Fig.~\ref{fig1}. In the figure, given the trained CNN models with~(without) fine-tuning, e.g. AlexNet~\cite{krizhevsky2012imagenet} and VGG Net~\cite{SimonyanZ14a}, we further extract the multi-scale CFVs, single-scale FCRs and our local discriminative MLRs for each image. Then the concatenated hybrid representations are fed into the linear SVM classifier~\cite{fan2008liblinear}. As validated by our experiments, the hybrid representations are very powerful features due to their complementary components, i.e., CFV which contains one or two order gradient information, FCR which contains global information of images, and MLR which contains local discriminative (structural) information. {\color{black}{The whole procedure of our hybrid representation is illustrated in Algorithm~\ref{Algorithm1}.}}

\subsection{MLR: Integrating Local Discrimination and CNN Features}
CNN has become a strong tool to learn invariant features for various visual applications. Nevertheless, during the training process of CNN, the input images are all global ones, and no good strategy has been found to incorporate the local information into the training process of CNN yet. In this paper, we propose a strategy to generate the mid-level local representation~(MLR), which is based on part dictionary clustering and multi-scale mid-level representation generating. MLR is constructed based on the local discriminative part dictionary, which leads to its local discrimination. In this subsection, we illustrate the details of our method for constructing the mid-level representation~(MLR) by utilizing the local discriminative ability of images.
\begin{figure}[t!]
	\centering
	\includegraphics[width=0.5\textwidth]{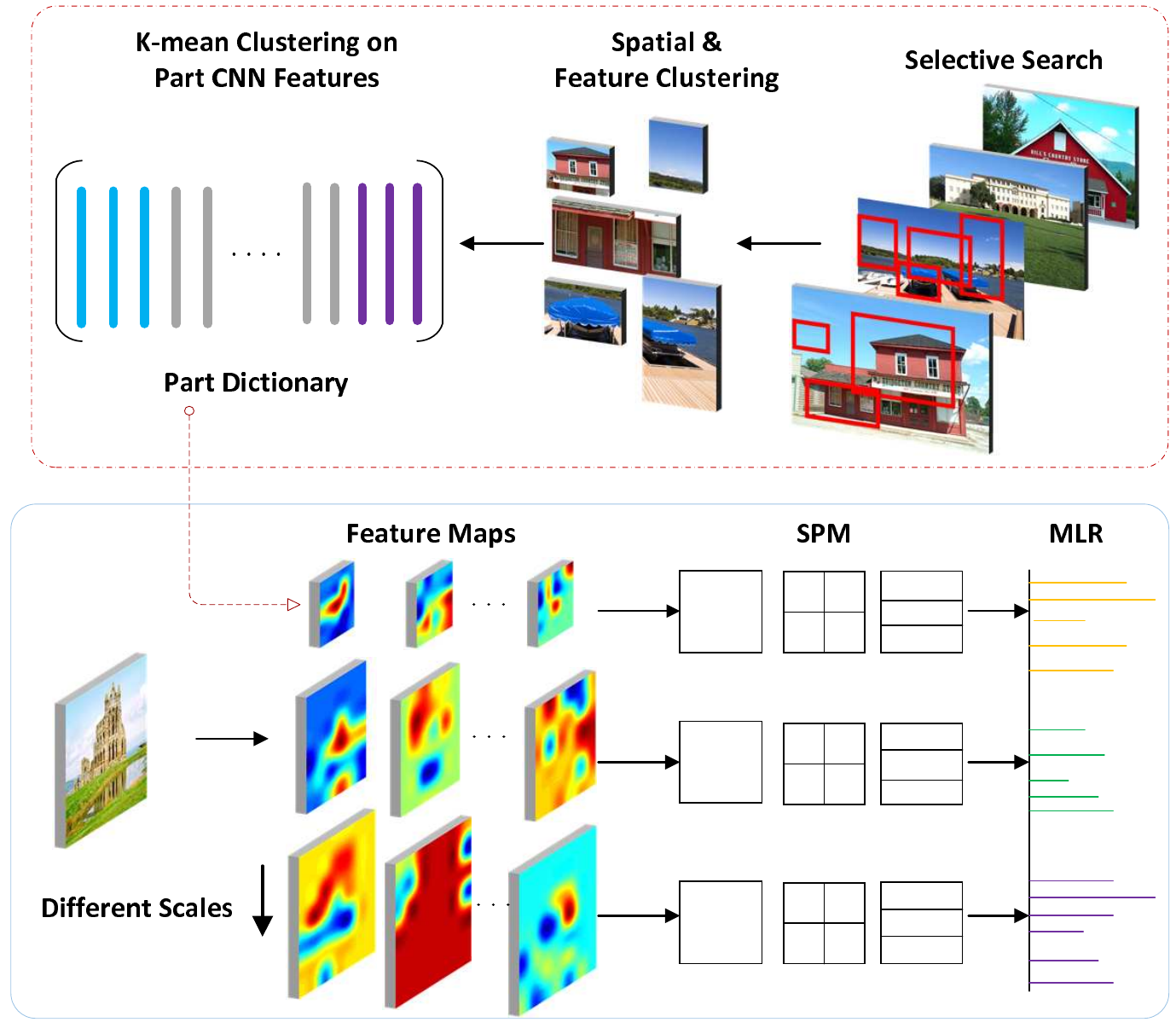}
	\caption{The process of calculating MLR based on the trained part dictionary. The operations in the red dashed box are the part dictionary generating process based on two-stage clustering, and operations in the below are the MLR generating process which is based on part dictionary and SPM. Best viewed in color.}
	\label{fig2}
\end{figure}

At the beginning, we utilize selective search~\cite{uijlings2013selective} to get the discriminative part proposals for every image, with the constraints of the pixel number of proposals within [60$\times$60,160$\times$160], and width/height~(height/width) smaller than 3, which is used to catch the local information of the images.

Furthermore, the spectral clustering in both the spatial and the feature space is conducted with the bounding box information of each image. Specifically, suppose the selected bounding boxes (by selective search) are $B = [B_{1},B_{2},\cdots,B_{n_{I}}]\in \mathbb{R}^{4\times n_{I}}$ with each $B_{i}$ denoting the coordinates of top-left and bottom-right on the image $I$, and the corresponding last fully connected layer activates of the trained CNN for $B$ are denoted as $F = [f_{1},f_{2},\cdots,f_{n_{I}}]\in \mathbb{R}^{d\times n_{I}}$~($d=4096$ for current popular networks). Herein, we use the activates after ReLU~\cite{krizhevsky2012imagenet}.  Under our constraints, while extracting $B$, the number of $n_{I}$ is usually less than 500, so the forward propagation for extracting $F$ will be acceptable. We construct the final similarity graph $G = (V,E)$ based on both $B$ and $F$, and the weights on the edges are as follows:
\begin{equation}
W = \lambda_{B}W_{B} + \lambda_{F}W_{F},
\label{Eq3}
\end{equation}
where $\lambda_{B}$ and $\lambda_{F}$ are the weighting parameters {\color{black}{and $\lambda_{B}+\lambda_{F}$=1 is used in our paper}} . Specifically, the elements of $W_{B}$ and $W_{F}$ are denoted as
\begin{equation}
W_{B}(i,j) = \dfrac{|B_{i}|\bigcap |B_{j}|}{|B_{i}|\bigcup |B_{j}|},
\label{Eq4}
\end{equation}
{\color{black}{where $|\cdotp|$ indicates calculating the area of the input box.}}
\begin{equation}
W_{F}(i,j) = \exp(-\dfrac{\|f_{i}-f_{j}\|^{2}_{2}}{2\sigma^{2}}),
\label{Eq5}
\end{equation}
for $i,j = 1,2,\cdots,n_{I}$.

{\color{black}{After spectral clustering on the graph $G$, we obtain $Q$ clusters $\mathcal{C}_{1},\mathcal{C}_{2},\cdots,\mathcal{C}_{Q}$. Then we sort the $Q$ clusters in descending order according to the number of bounding boxes and select the top $T$~($T\leq Q$) clusters.}} We randomly select one bounding box from each $\mathcal{C}_{i},i=1,2,\cdots,T$, thus totally $T$ bounding boxes are reserved. We further implement context padding~\cite{girshick2014rich} on the selected $T$ bounding boxes and do forward propagation (based on CNN) again to get the new feature representation based on the context padded boxes. In this way, we can get $T$ representations $P = [P_{1},P_{2},\cdots,P_{T}]$ for each image $I$, and we denote them as the prototypes of the corresponding image.

Finally, for all the prototypes of the training images, we do K-means clustering~\cite{kanungo2002efficient} to get the class-specific part dictionary and the class-mixture part dictionary, denoted as $D_{cs}\in \mathbb{R}^{d\times K}$ and $D_{cm}\in \mathbb{R}^{d\times K}$ respectively. Here the elements of $D_{cs}$ are obtained by first clustering the prototypes from each class and then concatenating them together. While $D_{cm}$ is based on clustering all the prototypes without considering the class information. The part dictionary contains local discriminative and multi-scale information of the training images, due to the selective search algorithm~(SSA) and constraints while running SSA.

With the part dictionary $D_{cs}$~($D_{cm}$) learned, we can consider it as a group of local discriminative filter banks. Motivated by Object-Bank~\cite{li2010object}, given an input image at a single scale, we sample square regions at multiple scales densely, i.e., sampled squares with size $128\times128$, $92\times92$ and $64\times64$ and step size 32 pixels for all three scales. Then we calculate the activates of the last fully connected layer of the trained CNN for all these squares under different scales, which generates three scaled activate tensors. This can also be seen as the local feature extraction on square regions. After that, for each scaled activate tensor of each image, we use the $D_{cs}$~($D_{cm}$) to operate with the activates on each location of the tensors, resulting in $K$ new feature maps for each scaled activate tensor, which can be seen as another kind of convolutional operation based on the part dictionary~\cite{kavukcuoglu2010learning}. For every $K$ feature maps under different scales, we further apply spatial pyramid matching (SPM)~\cite{lazebnik2006beyond} to divide the map region into spatial cells in three levels, i.e., $1\times1$, $2\times2$, $3\times1$, and do max-pooling on each cell. The final MLR is the concatenation of all the max-pooled features, with the dimension being $3\times K\times(1\times1+2\times2+3\times1)$.

Many methods can be adopted as the strategy of operation between the part dictionary $D_{cs}$~($D_{cm}$) and the activates of local square regions, e.g., inner production, sparse coding~\cite{yang2009linear}, locality-constrained linear coding~(LLC)~\cite{wang2010locality}, and auto-encoder based coding~\cite{Xie2014ACCV}. There are detailed coding speed comparisons in~\cite{Xie2014ACCV}. Here we utilize the LLC for feature coding, due to its relative fast speed and locality-preserve property. 

Specifically, denote the activating tensor of the image $I$ under one scale by $A\in \mathbb{R}^{d\times h \times h}$, and $A_{(\cdotp,i,j)}\in \mathbb{R}^{d}$, $(i,j = 1,2,\cdots,h)$ is the activating vector on the location $(i,j)$ of the $h\times h$ input tensor. To obtain the values $v^{\ast}\in \mathbb{R}^{K}$ on the location $(i,j)$ of $K$ new maps, we only need to solve the following LLC problem:
\begin{equation}
v^{\ast} = \arg\min_{v} \|A_{(\cdotp,i,j)}-Dv\|_{2}^{2}+\lambda\|dist\odot v\|_{2}^{2},
\label{Eq6}
\end{equation}
where $D$ can be taken as $D_{cs}$ or $D_{cm}$, $\odot$ denotes the element-wise multiplication, and $dist = [\exp(\|v-d_{1}\|^{2}/\tau),\exp(\|v-d_{2}\|_{2}^{2}/\tau),\cdots,\exp(\|v-d_{K}\|_{2}^{2}/\tau)]\in \mathbb{R}^{K}$ is the adaptive vector between $v$ and each dictionary element, which preserves the locality between $v$ and the dictionary $D$. In this paper, we utilize the approximated LLC~\cite{wang2010locality} for the fast calculation of $v^{\ast}$.

The flowchart of generating the MLR given the part dictionary $D_{cs}$ or $D_{cm}$ is shown in Fig.~\ref{fig2}.

\begin{algorithm}[t!]
	\scriptsize{
		\renewcommand{\algorithmicrequire}{\textbf{Input:}}
		\renewcommand\algorithmicensure {\textbf{Output:} }
		\caption{Extracting of Hybrid Representation}
		\small
		\label{alg2}
		\begin{algorithmic}[1]
			\REQUIRE
			{Training~(source) images: $\{I_{i}\}_{i = 1}^{m}$} and test~(target) images: $\{I_{i}\}_{i = m+1}^{n}$. A CNN model with or without fine-tuning.
			\ENSURE{Hybrid Representation $\{H_{i}\}_{i = 1}^{n}$ for each training and test images.}
			\renewcommand{\algorithmicrequire}{\textbf{Procedure:}}
			\REQUIRE
			\FOR{$i = 1\rightarrow n$}
			\STATE $\textbf{Selective search}$ on image $I_{i}$, obtain part set $B = [B_{1},B_{2},\cdots,B_{n_{I}}]\in \mathbb{R}^{4\times n_{I}}$. 
			\STATE $\textbf{Graph construction}$ based on Eq.~\ref{Eq3},~\ref{Eq4},~\ref{Eq5}, obtain graph $G$ with weight as $W$.
			\STATE $\textbf{Spectral clustering}$ on $G$, obtain $Q$ clusters.
			\STATE $\textbf{Sort}$ the $Q$ clusters by the number of boxes contained in each cluster.
			\STATE $\textbf{Select}$ the top $T$ clusters, and randomly take one box from each of the $T$ cluster.  
			\STATE $\textbf{K-means clustering}$ on the selected parts, generate class-specific or class-mixture dictionary. 
			\STATE $\textbf{LLC coding~(Eq.~\ref{Eq6}) and SPM}$, generate the $\textbf{MLR}_{i}$		
			\STATE $\textbf{Fisher vector coding}$~(Eq.~\ref{Eq2}) on the last convolutional layer of CNN, obtain $\textbf{CFV}_{i}$.
			
			\STATE $\textbf{Fully connected representations}$, $\textbf{FCR1}_{i}$ and $\textbf{FCR2}_{i}$ \\
			\STATE $H_{i} = [\textbf{MLR}_{i},\textbf{CFV}_{i},\textbf{FCR1}_{i},\textbf{FCR2}_{i}]$
			\ENDFOR
		\end{algorithmic}
		\label{Algorithm1}
	}
\end{algorithm}
\subsection{CFV: Convolutional Fisher Vector}
 In this part, we briefly describe the construction of the Fisher vector~\cite{perronnin2007fisher} based on the last convolutional layer of CNN. Like~\cite{perronnin2007fisher},~\cite{gong2014multi},~\cite{cimpoi2014deep}, and~\cite{yoo2014fisher}, we also use multiple scales as input while constructing CFVs.

 Moreover, considering the different scale information, we also calculate CFVs for each scale followed by $L2$ normalization and max-pooling. This strategy is known as Multi-scale Pyramid Pooling~\cite{yoo2014fisher}. The difference between our method and~\cite{yoo2014fisher} is that for sampling descriptors for GMM~\cite{reynolds2000speaker} training before Fisher vector coding, we also consider scale information (i.e., the number of sampled descriptors is proportional to that of the total descriptors under different scales in each image). Here the descriptors from the last convolutional layer are without ReLU~\cite{krizhevsky2012imagenet} throughout our paper.

 To improve the performance, power and $L2$ normalization are further applied to the max-pooled representation, which generates the final CFV representation.
 
\subsection{FCR: CNN Features from Well-Tuned Networks}
Given the well-trained CNN model~(with or without fine-tuning, such as AlexNet and VGG Net), we can use it for other different visual tasks, which is termed as parameter transfer learning. The fully connected representation (FCR) of the resized input images are extracted by forward propagating them until the fully connected layers of the network. We denote FCR1 and FCR2~(Fig.~\ref{fig1}) as the penultimate and the last fully connected layer representation with ReLU~\cite{krizhevsky2012imagenet}, respectively. Under this definition, our MLR is constructed based on FCR2 and local discriminative part (spectral clustering) on each image. 

{\color{black}{Before being fed into SVM classifier for training, $L$2 normalization is also applied to the FCRs.}}

\section{Experiments}
In this section, we show the classification accuracy of our hybrid representation in two applications, i.e., scene recognition and visual domain adaptation, compared with state-of-the-art models, including traditional and CNN based ones. We first introduce the datasets and experimental settings, then report experimental results on each dataset, after that we give an analysis of the key parameters followed by the analysis of the complementary ability of the proposed hybrid representation with the representations from other Nets, and finally we analysis the time complexity of calculating our representation. 

\subsection{Datasets and Experimental Settings}
For scene recognition, we utilize three trained CNN models, i.e., AlexNet~\cite{krizhevsky2012imagenet}, VGG-19 net~\cite{SimonyanZ14a}, GoogLeNet~\cite{szegedy2014going}, and VGG-11~\cite{WangGHQ15} net based on their caffe implementations~\cite{jia2013caffe}. Specifically, AlexNet and VGG-19 net are trained on ImageNet database, and these two nets are used to evaluate our hybrid representation system. GoogLeNet and VGG-11 net are trained on Place205 database, which are used to validate the complementary ability of our hybrid representation w.r.t. the representations from other nets. For domain adaptation, we use the AlexNet only, so that we can compare with other CNN based methods fairly. In our experiments, we resize all the images into resolution of $256\times256$ before the following operations, such as selective search and Fisher vector extraction. After obtaining the hybrid representation, we use linear SVM to train the classifier in all our experiments. The databases and experimental details are as follows:

\begin{figure}[t!]
	\centering
	\includegraphics[width=0.40\textwidth]{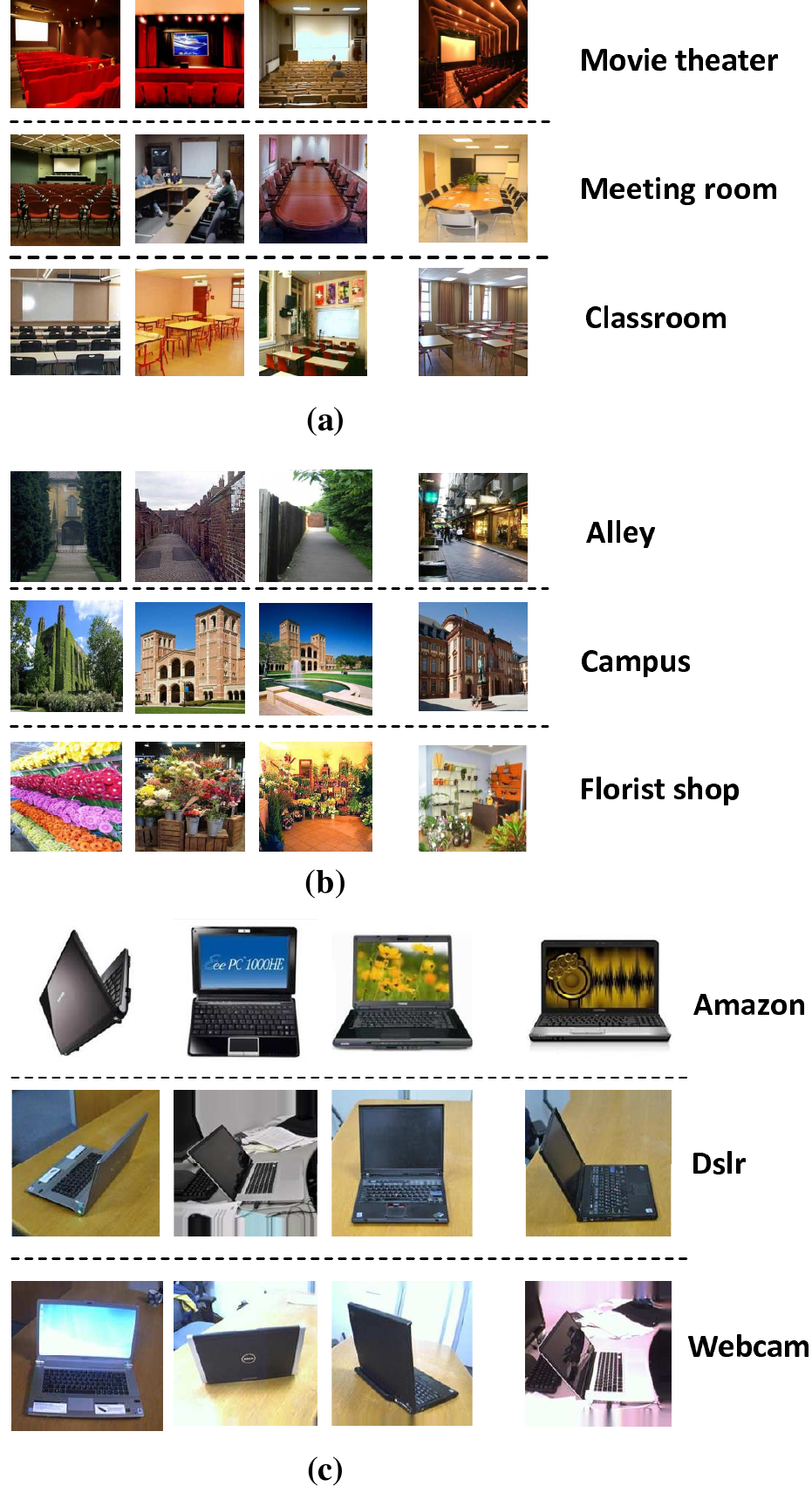}
	\caption{The sample images from different databases. (a) Images from MIT Indoor-67 database. It can be seen that categories ``movie theater", ``meeting room" and ``classroom" are very similar and difficult to distinguish. (b)~Images from SUN-397 database. It includes outdoor scenes, e.g. ``alley" and ``campus", and indoor scenes, e.g. ``florist shop". (c) Images from ``dest-top" category in Office database.  It contains 3 sub-sets, i.e., Amazon, Dslr, and Webcam. Best viewed in color.}
	\label{fig3}
\end{figure}

\subsubsection{$\textbf{MIT Indoor-67 database}$}
MIT Indoor-67 database~\cite{quattoni2009recognizing} is a popular indoor scene database, including 15,620 images of 67 indoor scenes. It is difficult to distinguish different classes, because the categories are all indoor scenes, and  inter-class variance between different classes is very little. We follow the same training-test partition as~\cite{quattoni2009recognizing} by using approximately 80 images from each class for training and 20 for testing. The average class accuracy is reported in our paper. Sample images of Indoor-67 can be found in Fig.~\ref{fig3}~(a). It can be seen from Fig.~\ref{fig3}~(a) that the categories of mover theater, meeting room and classroom are very similar.
\subsubsection{$\textbf{SUN-397 database}$}
SUN-397~\cite{xiao2010sun} is a large-scale scene recognition database, containing 397 categories and with scenes varying from abbey to zoo, including both indoor and outdoor scenes. At least 100 images are contained in each category. We use the publicly available training-test partitions, and report the average accuracy and standard errors for all the partitions. We use 50 images from each class for training and 50 for testing. Sample images can be found in Fig.~\ref{fig3}~(b).
\subsubsection{$\textbf{Office database}$}
Office dataset~\cite{saenko2010adapting} contains three sub-datasets from different domains, i.e., Amazon, Webcam,  and Dslr. Each sub-dataset is from a separate domain: images from Amazon are collected from online catalogs of amazon.com, and images in Dslr and Webcam are obtained in the daily office environment by a digital SLR camera and a webcam with high and low resolutions, respectively. There are totally 31 categories which are common for these three domains, and the number of images per category per domain ranges from 8 to 100, and totally 4,652 images are included. Fig.~\ref{fig3}~(c) shows the domain shift between different domains.
\subsubsection{$\textbf{Details for domain adaptation}$}
In domain adaptation, if the training data~(source domain) with labels and the test data~(target domain) without labels are given, it is called unsupervised domain adaptation; if a source domain with labels and a target domain with a small amount of labeled data are given, then the problem is denoted as semi-supervised domain adaptation. We will perform both unsupervised and semi-supervised domain adaptation on all the 31 categories in our experiments. We adopt the standard experimental setup presented in~\cite{saenko2010adapting}. We use 20 source examples per category when Amazon is taken as  the source domain, and 8 images per category when Webcam or Dslr is taken as the source domain~\cite{saenko2010adapting}. For the semi-supervised adaptation, three more labeled target examples per category are added into the source domain. We also try another setting presented in~\cite{gong2013connecting},~\cite{Menglong2015deep}, where we use all the source domain examples with labels for unsupervised adaptation and three more target domain samples per class for semi-supervised adaptation.

We evaluate our method across five random training-test partitions for each of the three domain transfer tasks commonly used for evaluation, i.e., Amazon$\rightarrow$Webcam~($A\rightarrow W$), Dslr$\rightarrow$Webcam~($D\rightarrow W$) and Webcam$\rightarrow$Dslr~($W\rightarrow D$), and report average accuracy and standard errors for each setting.

\begin{figure*}[t!]
	\centering
	\includegraphics[width=0.85\textwidth]{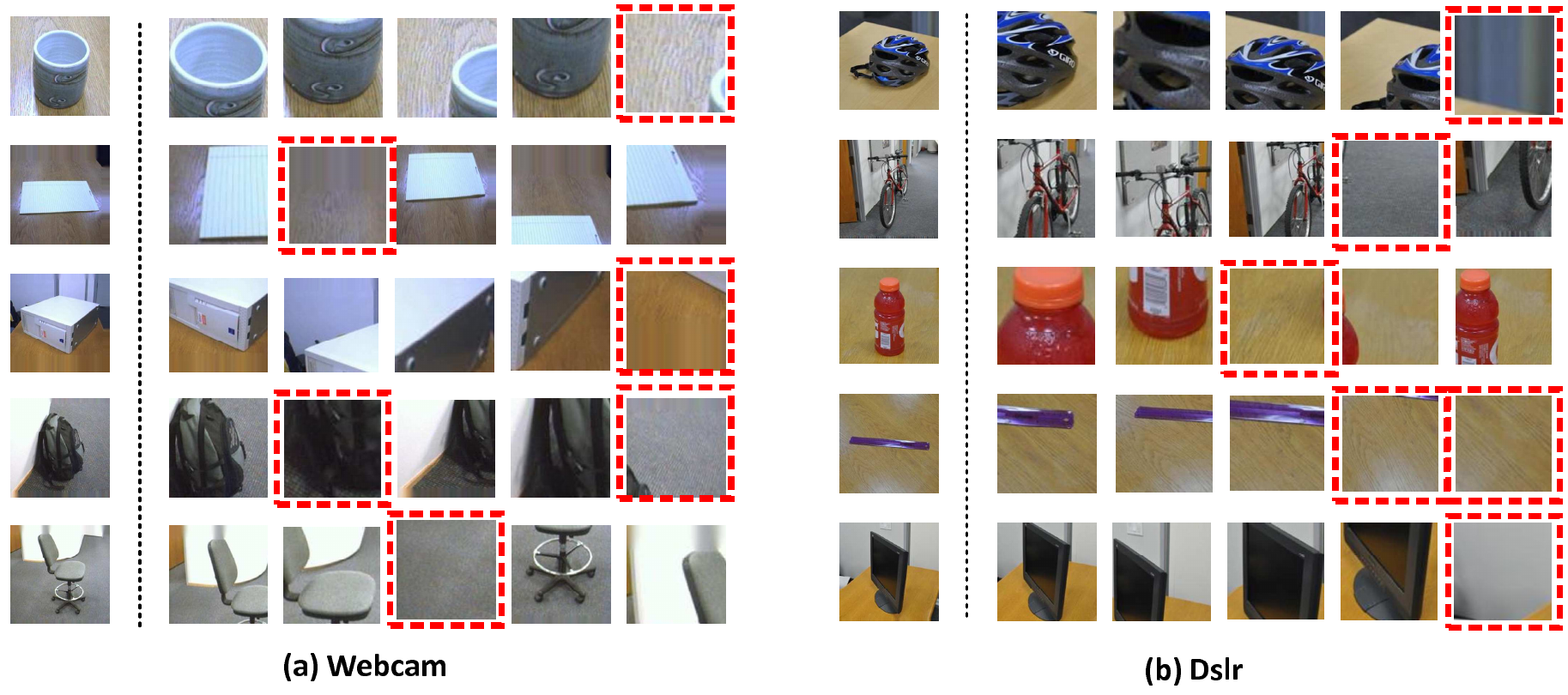}
	\caption{Examples of original images~(first column in both (a) Webcam and (b) Dslr domains) and their corresponding prototype boxes~(2nd-6th columns) based on our proposed first stage spectral clustering. The boxes are resized to fit CNN input. Red dashed boxes are removed ones based on Eq.~\ref{Eq7}.  Best viewed in color.}
	\label{fig4}
	\vspace{-0.5cm}
\end{figure*}

{\color{black}{Actually, given the CNN model, calculating the hybrid representation based on Algorithm~\ref{Algorithm1} is already a domain~(parameter) transfer process. As for domain adaptation experiments, we run Algorithm~\ref{Algorithm1} to generate hybrid representation for both source and target data, which is used for further domain transfer by training classifier on the source representation, and testing on the target one.}} Note that in domain adaptation, after obtaining the prototype boxes based on the first stage spectral clustering, we first carry out box filtering based on variance thresholding of the prototype boxes, followed by the second stage K-means clustering. The purpose of this trick is to filter out the box regions with low variance, which may be the surrounding background regions in the Office database. Specifically, given the prototype box~$I_{b}$, we calculate the variance of its gray scale image as follows:
\begin{equation}
VAR = var(gray(I_{b})).
\label{Eq7}
\end{equation}
Then if $VAR < 125$, we delete the prototype. Fig.~\ref{fig4} illustrates the deleted prototype boxes based on our strategy.
\subsubsection{$\textbf{Parameter settings}$} 
{\color{black}In Algorithm~\ref{Algorithm1}, there are several key parameters, we give a detailed description of these parameters. As for graph constructing, the parameter $(\lambda_{B},\lambda_{F})$ in~Eq.~\ref{Eq3} is set as $(1,0)$ for Indoor-67 database, and $(0.5,0.5)$ for SUN397 and Office databases~\footnote{ Without considering the feature space information on Indoor-67~($\lambda_{F}=0$) is based on the observations: the layout in Images of Indoor-67 is almost dense, while the layouts in images of SUN-397 and Office are sparse, i.e., some sub-regions are the same, e.g. the sky in the campus category of SUN-397, and the background of ruler category in Office dataset.}. $\sigma$ in Eq.~\ref{Eq5} is set as $1$. The number $G$ of clusters for spectral clustering is set as $10$, and the number of top ranking $T$ clusters is set as $5$, which is equal to the number of boxes selected from each image. As for Fisher vector coding, the Gaussian components of GMM training is fixed as $64$, and we use multiple scales in our paper. Specifically, given the CNN input size of  $L\times L$ for the image $I$, the used five scales are $L \times \sqrt{2}^{[0\, 1\, 2\, 3\, 4]} = [L\; \sqrt{2}L\; 2L\; 2\sqrt{2}L\; 4L]$~\footnote{Note that for Indoor-67 experiment under VGG-19 net, we use the 10 scales the same as~\cite{cimpoi2014deep}, i.e., $L \times \sqrt{2}^{[-6\,-5\,-4\,-3\,-2\,-1\,0\, 1\, 2\, 3]}$   to reproduce their results.}. As for FCR1 and FCR2, we extract features with the whole image as input for Indoor-67 and SUN-397, and extract features with the central cropped sub-region as input for Office. The SVM parameter $C$ is fixed as $1$ in all our experiments.}
\subsubsection{$\textbf{Abbreviation in the experiments}$} 
To make some key definition clear, we list these abbreviation in Table~\ref{table1}.
 \begin{table}[!t]
 	\renewcommand{\arraystretch}{1}
 	\caption{Abbreviation in this paper.}
 	\label{table_example}
 	\centering
 	\scriptsize{\begin{tabular}{|c|c|}
 		\hline
 		\textbf{Abbreviation}& Description\\
 		\hline
 		FCR1-c & FCR1 based on central crop as input  \\
 		\hline
 		FCR1-w & FCR1 based on whole image as input  \\
 		\hline
  		FCR2-c & FCR2 based on central crop as input  \\
  		\hline
  		FCR2-w & FCR2 based on whole image as input  \\
  		\hline
  		CM & Construct MLR based on class-mixture dictionary  \\
  		\hline		
  		CS & Construct MLR based on class-specific dictionary  \\
  		\hline 
  		G\_P205 & $\begin{aligned}
  		\text{The average pooled representation before loss3} \\
  		\text{of GoogLeNet~(trained on Place205)}
  		\end{aligned}$ \\
  		\hline
  		VGG-11\_P205 & FCR1-w of VGG-11 trained on Place205   \\
  		\hline 
  		VGG-19\_Hybrid & FCR1-w+CFV+MLR based on VGG-19 trained on ImageNet  \\
  		\hline
 	\end{tabular}}
 		\vspace{-0.5cm}
 		\label{table1}
 \end{table}

\begin{table}[!t]
	\renewcommand{\arraystretch}{1}
	\caption{Comparison of classification rate on MIT Indoor-67 database.}
	\label{table_example}
	\centering
	\scriptsize{\begin{tabular}{|c|c|c|}
			\hline
			\textbf{Traditional Methods} & \multicolumn{2}{c|}{Accuracy~(\%)} \\
			\hline
			\hline
			ROI~\cite{quattoni2009recognizing} & \multicolumn{2}{c|}{26.05}\\
			\hline
			MM-Scene~\cite{zhu2010large} & \multicolumn{2}{c|}{28.00}\\
			\hline
			DPM~\cite{pandey2011scene} &  \multicolumn{2}{c|}{30.40}\\
			\hline
			CENTRIST~\cite{wu2011centrist} &  \multicolumn{2}{c|}{36.90}\\
			\hline
			Object Bank~\cite{li2010object} &  \multicolumn{2}{c|}{37.60}\\
			\hline
			RBOW~\cite{parizi2012reconfigurable} &  \multicolumn{2}{c|}{37.93}\\
			\hline
			Discriminative Patches~\cite{singh2012unsupervised} &  \multicolumn{2}{c|}{38.10} \\
			\hline
			Hybrid parts~\cite{zheng2012learning} &  \multicolumn{2}{c|}{39.80} \\
			\hline
			LPR-LIN~\cite{sadeghi2012latent} &  \multicolumn{2}{c|}{44.84}\\
			\hline
			BOP~\cite{juneja2013blocks} &  \multicolumn{2}{c|}{46.10} \\
			\hline
			MI-SVM~\cite{li2013harvesting} &  \multicolumn{2}{c|}{46.40} \\
			\hline
			Hybrid parts+GIST-color+SP~\cite{zheng2012learning} &  \multicolumn{2}{c|}{47.20} \\
			\hline
			ISPR~\cite{lin2014learning} &  \multicolumn{2}{c|}{50.10} \\
			\hline
			MMDL~\cite{wang2013max} &  \multicolumn{2}{c|}{50.15} \\
			\hline
			Discriminative Parts~\cite{sun2013learning} &  \multicolumn{2}{c|}{51.40}\\
			\hline
			DSFL~\cite{zuo2014learning} &  \multicolumn{2}{c|}{52.24} \\
			\hline
			Discriminative Lie Group~\cite{ludiscriminative} &  \multicolumn{2}{c|}{55.58} \\
			\hline
			IFV~\cite{juneja2013blocks} &  \multicolumn{2}{c|}{60.77} \\
			\hline
			IFV+BOP~\cite{juneja2013blocks} &  \multicolumn{2}{c|}{63.10} \\
			\hline
			Mode-Seeking~\cite{doersch2013mid} &  \multicolumn{2}{c|}{64.03} \\
			\hline
			Mode-Seeking + IFV~\cite{doersch2013mid} &  \multicolumn{2}{c|}{66.87} \\
			\hline
			ISPR+IFV~\cite{lin2014learning} &  \multicolumn{2}{c|}{68.50} \\
			\hline
			\hline
			\textbf{CNN based Methods} &  \multicolumn{2}{c|}{Accuracy~(\%)}\\
			\hline
			\hline
			FCR2~(placeNet)~\cite{zhou2014learning} &  \multicolumn{2}{c|}{68.24} \\
			\hline
			Order-less Pooling~\cite{gong2014multi} &  \multicolumn{2}{c|}{68.90} \\
			\hline
			CNNaug-SVM~\cite{razavian2014cnn} &  \multicolumn{2}{c|}{69.00} \\
			\hline
			FCR2~HybridNet~\cite{zhou2014learning} &  \multicolumn{2}{c|}{70.80} \\
			\hline
			URDL+CNNaug~\cite{Liu2014ACCV} &  \multicolumn{2}{c|}{71.90} \\
			\hline
			MPP-FCR2~(7 scales)~\cite{yoo2014fisher} &  \multicolumn{2}{c|}{75.67} \\
			\hline
			DSFL+CNN~\cite{zuo2014learning} &  \multicolumn{2}{c|}{76.23} \\
			\hline
			MPP+DSFL~\cite{yoo2014fisher} &  \multicolumn{2}{c|}{80.78} \\
			\hline
			CFV~(VGG-19)~\cite{cimpoi2014deep} &  \multicolumn{2}{c|}{81.00} \\
			\hline
			\hline
			\multirow{2}{*}{\textbf{Our hybrid Methods~(CS)}} & \multicolumn{2}{c|}{Accuracy~( \%)} \\ 
			\cline{2-3}
			& AlexNet & VGG-19\\
			\hline
			\hline
			FCR1-c & 59.71 & 72.96\\
			\hline
			FCR1-w & 61.63 & 71.48\\
			\hline
			FCR2-c & 59.48 & 70.23\\
			\hline
			FCR2-w & 61.01 & 68.71\\
			\hline
			FCR2-w~(fine-tuning) & 64.78 & $-$\\
			\hline
			MLR & 69.33 & 77.51\\
			\hline
			CFV & 68.68 & 81.05\\
			\hline
			FCR1-w+FCR2-w & 62.43 & 70.59\\
			\hline
			MLR+FCR2-w & 70.20 & 77.94 \\
			\hline
			CFV+FCR2-w & 71.27 & 80.75\\
			\hline
			MLR+FCR1-w & 70.66 & 78.81\\
			\hline
			CFV+FCR1-w & 70.94 & 81.60\\
			\hline
			MLR+CFV & 72.80 & 81.47 \\
			\hline
			MLR+CFV+FCR1-w & $\textbf{74.09}$ & $ \textbf{82.24}$\\
			\hline
			MLR+CFV+FCR2-w & 73.17 & 81.57 \\
			\hline
			CFV+FCR1-w+FCR2-w & 70.72 & 79.70 \\
			\hline
			MLR+FCR1-w+FCR2-w & 69.57 & 78.34  \\
			\hline 
			MLR+CFV+FCR1-w+FCR2-w & 72.98 & 80.91  \\
			\hline     
		\end{tabular}}
		\vspace{-0.5cm}
        \label{table2}
	\end{table}

\subsection{MIT Indoor-67 Experiments}

\begin{figure}[t!]
	\centering
	\includegraphics[width=0.5\textwidth]{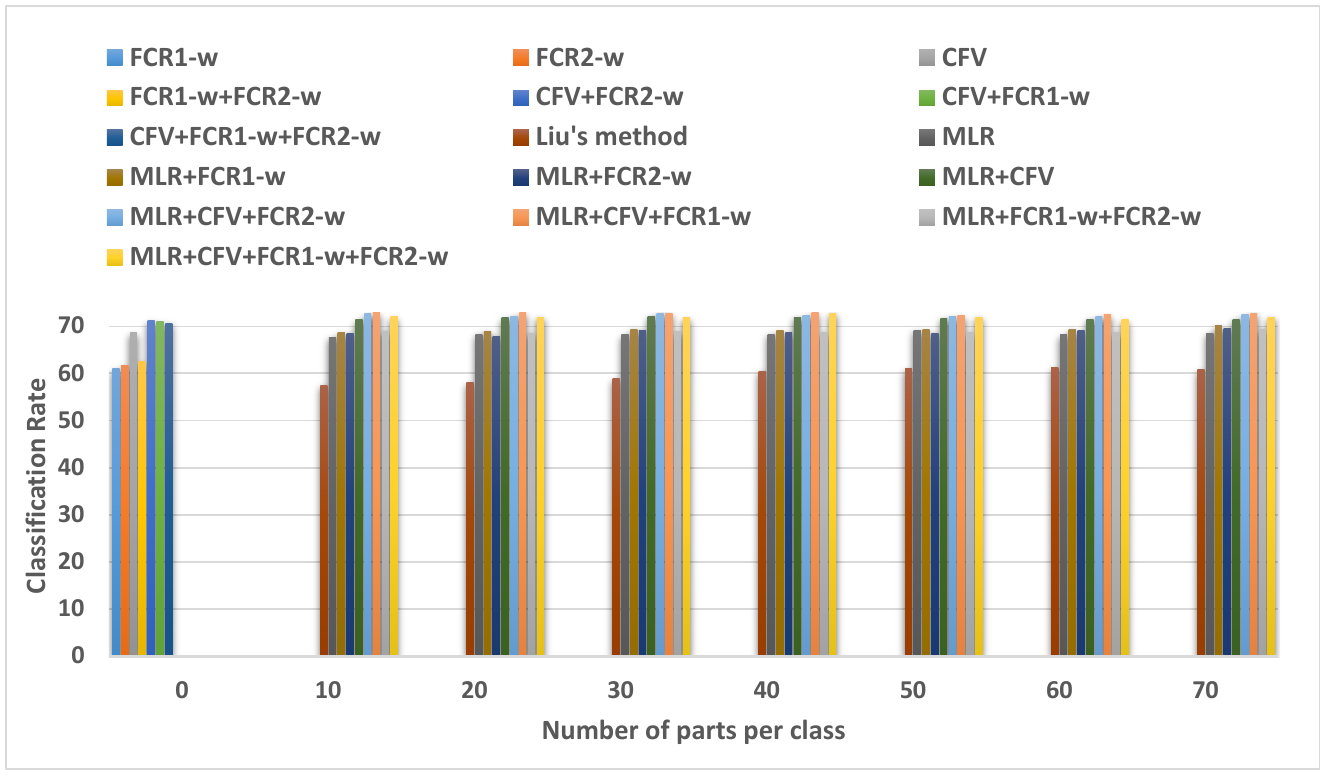}
	\caption{Changing tendency of classification rate with different part element numbers of class-mixture dictionary $D_{cm}$ on Indoor-67 database, based on AlexNet. Best viewed in color.}
	\label{fig5}
\end{figure}

\begin{figure}[t!]
	\centering
	\includegraphics[width=0.5\textwidth]{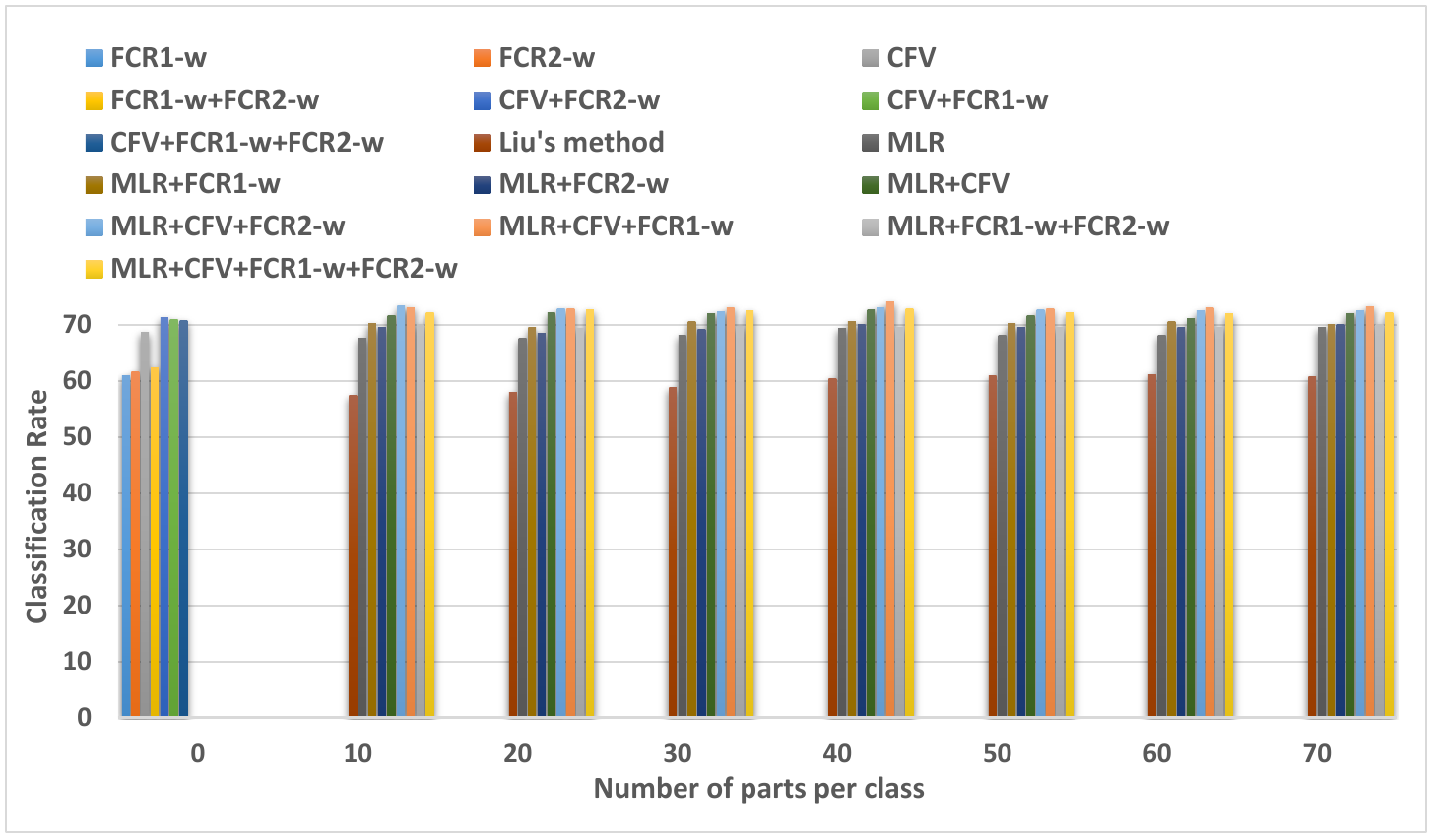}
	\caption{Changing tendency of classification rate with different part element numbers of class-specific dictionary $D_{cs}$ on Indoor-67 database, based on AlexNet. Best viewed in color.}
	\label{fig6}
	\vspace{-0.5cm}
\end{figure}

\begin{figure}[t!]
	\centering
	\includegraphics[width=0.5\textwidth]{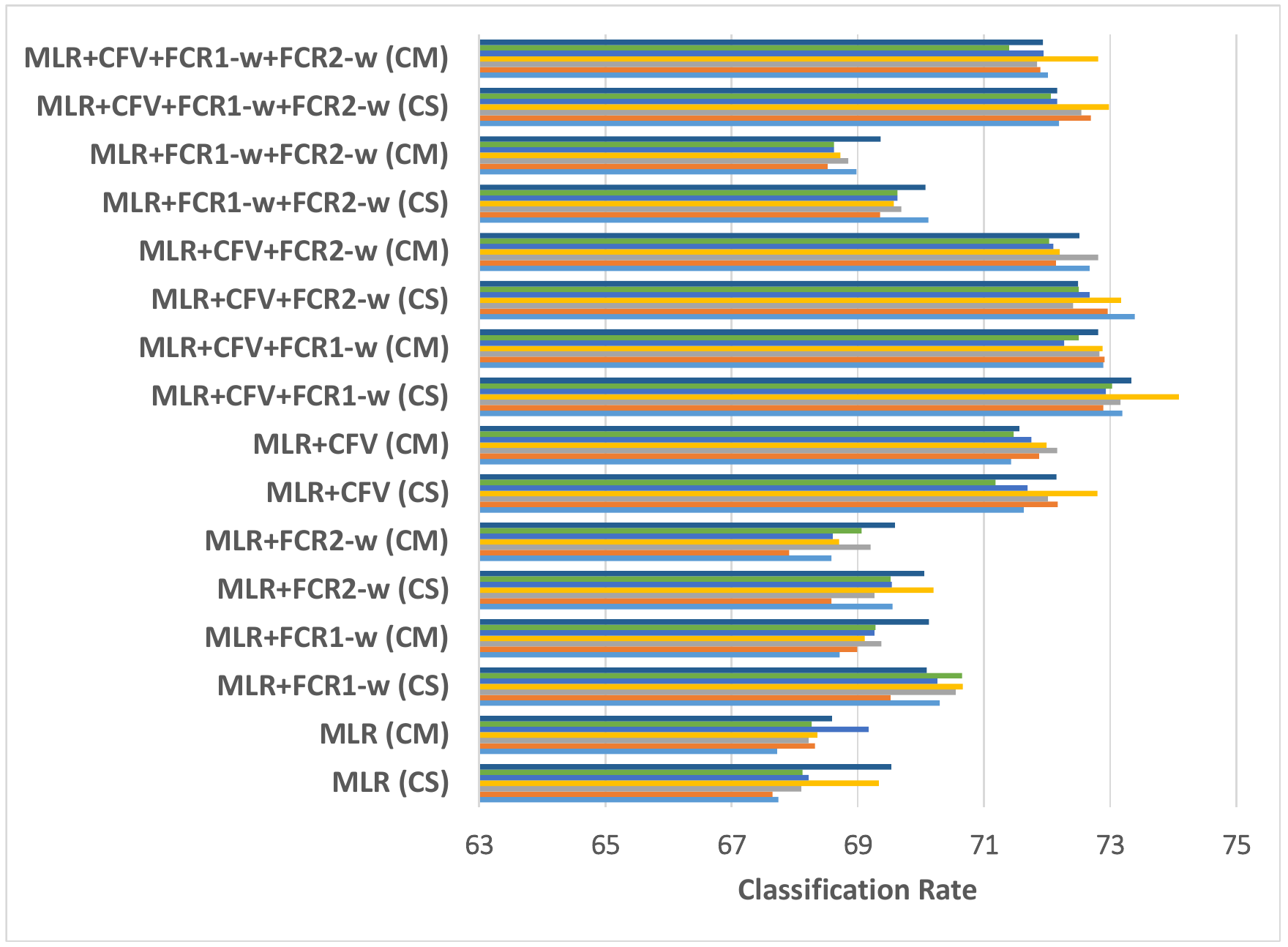}
	\caption{Comparison of changing tendency of classification rate under different part element numbers and different part dictionaries, i.e., Class-Mixture~(CM) and Class-Specific~(CS) on Indoor-67 database, based on AlexNet. Take MLR(CS) as an example. The accuracies from bottom to top are corresponding to the part element number that varies from $10\times67$ to $70\times67$. Best viewed in color.}
	\label{fig7}
		\vspace{-0.5cm}
\end{figure}

In this Subsection, we report and analyze the experimental results on the MIT Indoor-67 database. We first show the changing tendency of the classification rate under different part dictionary sizes, in the class-mixture and the class-specific manner respectively, without fine-tuning the CNN. The part dictionary size can be seen as the number of representative parts per category multiplying the total category number (see Fig.~\ref{fig5}-\ref{fig7} for details).

In both Fig.~\ref{fig5} and Fig.~\ref{fig6}, ``0" on the x-axis means that representations FCR1-w, FCR2-w, CFV, FCR1-w+FCR2-w, CFV+FCR2-w, CFV+FCR1-w and CFV+FCR1-w+FCR2-w are not generated based on the part dictionary. The detailed meaning of FCR1-w and FCR2-w are listed in Table~\ref{table1}. From Fig.~\ref{fig5} and~\ref{fig6}, it can be seen that our part dictionary is much better than that of Liu et al.~\cite{Liu2014ACCV} in the classification rate under different part dictionary sizes, and the hybrid representations combined with MLR are much better than their counterpart representations. From Fig.~\ref{fig7},  a conclusion can be drawn that class-specific dictionaries are always better than class-mixture ones. Moreover, learning a class-specific dictionary is much faster than learning a class-mixture one based on K-means, thus the best choice of the second stage clustering is the class-specific part dictionary.

To better evaluate the representation ability of our proposed MLR, we also compare the classification rate with or without fine-tuning based on AlexNet~\cite{krizhevsky2012imagenet}. We use the AlexNet with the architecture of Caffe's implementation~\cite{jia2013caffe}, which contains five convolutional layers, two fully connected layers and one output layer with a node number equal to the number of categories (i.e., the output number of nodes is $67$ for Indoor-67 database). We first fine-tune AlexNet with all the global training images~(resized to $256\times 256$) of Indoor-67. The initialization of the parameters of the first seven layers is the same as the model trained on ImageNet database, and the parameters of the last layer are randomly initialized with Gaussian distribution. The learning rates of the first seven layers and the last fully-connected layer are initialized as
0.001 and 0.01 respectively, and reduced to one tenth of the current rates after fixed iterations~(10,000 in our experiments). By setting a larger learning rate for the last layer, due to the random initialization of this layer, we hope the parameters of this layer can be updated to be more suitable for the new task-specific database. For the previous layers of AlexNet, we hope the parameters change as little as possible to preserve the already learned texture and shape information during the learning based on ImageNet database. Then based on our fine-tuned task-specific model, we further fine-tune the AlexNet based on the prototype bounding boxes~(with labels the same as the images where they are located on), which are extracted based on the first stage spatial and feature spectral clustering. 
  
\begin{figure}[t!]
	\centering
	\includegraphics[width=0.5\textwidth]{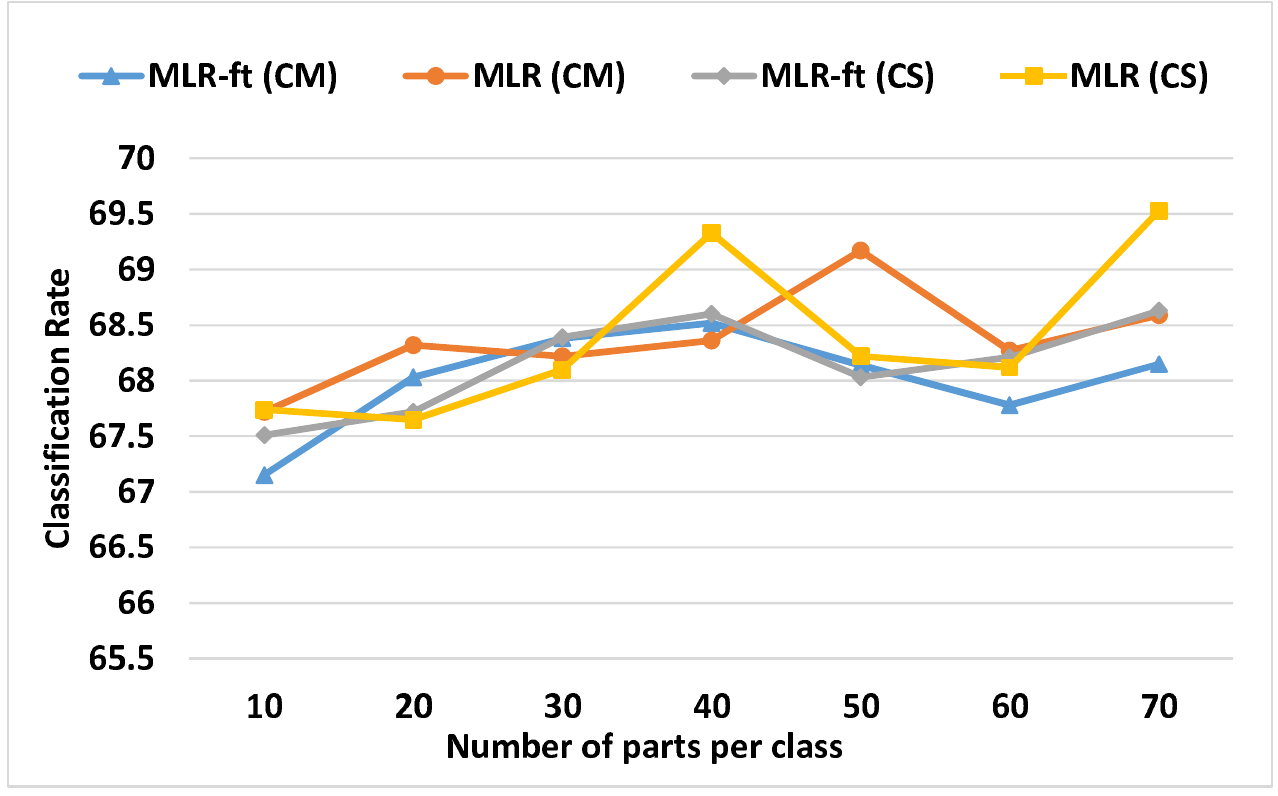}
	\caption{Comparison of classification rate~(with or without fine-tuning) under different part element numbers and different dictionaries on Indoor-67 database, based on AlexNet. Best viewed in color.}
	\label{fig8}
		\vspace{-0.5cm}
\end{figure}

An observation is that the classification rate of MLR based on fine-tuned AlexNet is comparable with their counterpart representation calculated based on AlexNet without fine-tuning. See Fig.~\ref{fig8} for details. Therefore, we do not fine-tune our used CNN models in the later experiments.

Finally, we list the classification accuracies on Indoor-67 and compare our hybrid methods with other state-of-the-art methods, both under AlexNet and VGG-19 networks. Table~\ref{table2} gives the detailed accuracies of the traditional models, the CNN based models, and our hybrid representation based methods. Here the dictionary size for constructing MLR is 67$\times$40 and 67$\times$30 for AlexNet and VGG-19, respectively. It can be seen that our hybrid representation achieves the result of 82.24$\%$ based on VGG-19 network, which has been the best result on Indoor-67 based on the net trained on ImageNet database. This justifies the effectiveness of combining CNN and dictionary-based features in improving the classification accuracy.
\begin{table}[!t]
	\renewcommand{\arraystretch}{1}
	\caption{Comparisons of classification rate on SUN-397 database.}
	\label{table_example}
	\centering
	\scriptsize{\begin{tabular}{|c|c|c|}
			\hline
			\textbf{Traditional Methods}&  \multicolumn{2}{c|}{Accuracy~(\%)}\\
			\hline
			\hline
			S-manifold~\cite{kwitt2012scene} &  \multicolumn{2}{c|}{28.90} \\
			\hline
			OTC~\cite{margolin2014otc} &  \multicolumn{2}{c|}{34.56} \\
			\hline
			contextBow+semantic~\cite{su2012improving} &  \multicolumn{2}{c|}{35.60} \\
			\hline
			Xiao et al.~\cite{xiao2010sun} &  \multicolumn{2}{c|}{38.00}\\
			\hline
			FV~(SIFT + Local Color Statistic)~\cite{sanchez2013image} &  \multicolumn{2}{c|}{47.20}\\
			\hline
			OTC + HOG2x2~\cite{margolin2014otc} &  \multicolumn{2}{c|}{49.60} \\
			\hline
			\hline
			\textbf{CNN based Methods} &  \multicolumn{2}{c|}{Accuracy~(\%)}\\
			\hline
			\hline
			Decaf~\cite{donahue2013decaf} &  \multicolumn{2}{c|}{40.94} \\
			\hline
			MOP-CNN~\cite{gong2014multi} &  \multicolumn{2}{c|}{51.98} \\
			\hline
			Hybrid-CNN~\cite{zhou2014learning} &  \multicolumn{2}{c|}{53.86$\pm$0.21}\\
			\hline
			Place-CNN~\cite{zhou2014learning} &  \multicolumn{2}{c|}{54.32$\pm$0.14}\\
			\hline
			Place-CNN ft~\cite{zhou2014learning} &  \multicolumn{2}{c|}{56.2}\\
			\hline
			\hline
			\multirow{2}{*}{\textbf{Our hybrid Methods~(CS)}} & \multicolumn{2}{c|}{Accuracy~( \%)} \\ 
			\cline{2-3}
			& AlexNet & VGG-19\\
			\hline
			\hline
			FCR2-w & 46.27$\pm$0.37 & 52.78$\pm$0.25\\
			\hline
			FCR1-w & 46.42$\pm$0.47 & 54.98$\pm$0.10 \\
			\hline
			MLR & 53.84$\pm$0.16 & 61.09$\pm$0.11 \\
			\hline
			CFV & 52.30$\pm$0.09 & 62.47$\pm$0.41 \\
			\hline
			FCR1-w+FCR2-w & 47.19$\pm$0.53 & 54.51$\pm$0.14  \\
			\hline
			MLR+FCR2-w &  55.22$\pm$0.34 & 62.09$\pm$0.24 \\
			\hline
			CFV+FCR2-w & 54.65$\pm$0.40 & 62.77$\pm$0.21 \\
			\hline
			MLR+FCR1-w &  55.17$\pm$0.33 & 62.59$\pm$0.14 \\
			\hline
			CFV+FCR1-w & 54.34$\pm$0.22 & 63.16$\pm$0.29 \\
			\hline
			MLR+CFV & 56.66$\pm$0.17 & 64.14$\pm$0.32 \\
			\hline
			MLR+CFV+FCR1-w & 57.15$\pm$0.26& $\textbf{64.53}$$\pm$$\textbf{0.24}$ \\
			\hline
			MLR+CFV+FCR2-w &$\textbf{57.31}$$\pm$$\textbf{0.12}$& 64.35$\pm$0.19 \\
			\hline
			CFV+FCR1-w+FCR2-w & 53.80$\pm$0.53 & 61.90$\pm$0.33\\
			\hline
			MLR+FCR1-w+FCR2-w & 54.84$\pm$0.32 & 62.05$\pm$0.17 \\
			\hline
			MLR+CFV+FCR1-w+FCR2-w & 56.83$\pm$0.13 & 64.13$\pm$0.16 \\
			\hline
		\end{tabular}}
		\label{table3}
		 	\vspace{-0.5cm}
	\end{table}
\subsection{SUN-397 Experiments}
In this subsection, we report and analyze the classification results on the SUN-397 database. It can be seen from Table~\ref{table3} that our hybrid representation can also achieve the best results under both AlexNet and VGG-19 net configurations based on the net trained on ImageNet database. In our experiments on SUN-397 database, to improve the computation efficiency, we fix the part number of each class as $10$ while generating the  class-mixture part dictionary $D_{cm}$ or the class-specific one $D_{cs}$ (i.e., the total number of elements for $D_{cm}$ and $D_{cs}$ is both 3790).

\begin{figure}[t!]
	\centering
	\includegraphics[width=0.5\textwidth]{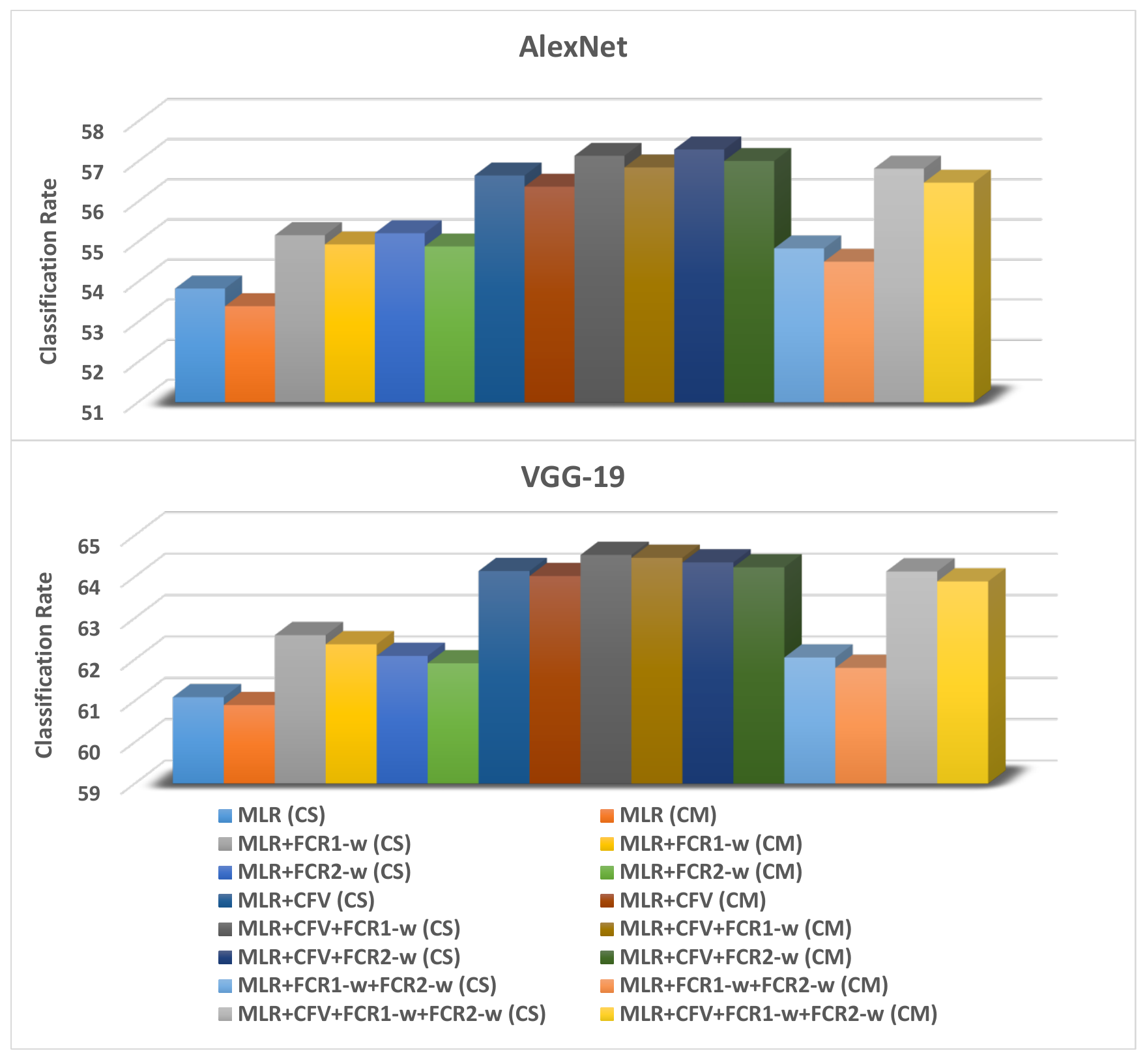}
	\caption{Comparison of classification rate under different part dictionaries~(dictionary size is fixed as 3970), i.e., Class-Mixture~(CM) and Class-Specific~(CS) on SUN-397 database, based on AlexNet and VGG-19 Net. Best viewed in color.}
	\label{fig9}
		\vspace{-0.5cm}
\end{figure}

Note that only under AlexNet, our best result is 57.31$\%$, which is already better than the current best result of 56.20$\%$ by fine-tuning the PlaceNet trained on the large place database~\cite{zhou2014learning}. On the other hand, we fix the part dictionary size as $3970$ in our experiments on this database. If we enlarge the dictionary size, better results can be further obtained. On this database, the performance of the class-specific dictionary is also better than the class-mixture counterpart (see Fig.~\ref{fig9} for details).

\subsection{Domain Adaptation Experiments}
In this part, we report and analyze the domain adaptation~(DA) experiments, under both unsupervised setting and semi-supervised setting with two different source-target partitions. The datasets used here are the Amazon, Webcam and Dslr datasets. We only conduct experiments based on AlexNet in DA experiments, which almost all yield best results under the four settings. If our representation is constructed under more strong CNN models, better improvements can be further obtained.

Table~\ref{table4}-\ref{table7} show the DA performances of our method compared with other methods. It can be seen that our results are much better than other state-of-the-art DA approaches under $D\rightarrow W$ and $W\rightarrow D$. For the domain transfer $A\rightarrow W$, only comparative results are obtained. Here, the reason may be that the domain biases between domains $A$ and $W$ are very large, which can greatly reduce the importance of local information. However, we can construct the hybrid representation based on the trained models of DDC~\cite{tzeng2014deep} and DAN~\cite{Menglong2015deep} which are better than our results based on AlexNet. Another strategy to enhance the representation ability of our representation is to use stronger CNN models, e.g., VGG-19 Net, to construct the hybrid representation. Surprisingly, we get $100\%$ domain transfer accuracy for transferring from Webcam to Dslr, under the setting of using all Webcam images and three more Dslr samples per class for training. This indicates that relatively small domain shifts can be totally avoided by our local representation~(MLR) combined with global representations~(FCR1-c+FCR2-c) of CNN features.

\begin{table}[!t]
	\renewcommand{\arraystretch}{1}
	\caption{Classification accuracy of 31-category Office dataset under unsupervised adaptation settings. 20 samples per class are used with Amazon as the source domain, and 8 samples per class with Webcam and Dslr as the source domain respectively.}
	\label{table_example}
	\centering
	\scriptsize{\begin{tabular}{|c|c|c|c|}
		\hline
		\textbf{Methods}& $A\rightarrow W$ & $D\rightarrow W$ & $W\rightarrow D$\\
		\hline
		\hline
		GFK~(PLS,PCA)~\cite{gong2012geodesic}  & 15.0$\pm$0.4& 44.6$\pm$0.3 & 49.7$\pm$0.5 \\
		\hline
		SA~\cite{fernando2013unsupervised}  & 15.3& 50.1 & 56.9 \\
		\hline
		DaNBNN~\cite{tommasi2013frustratingly}  & 23.3$\pm$2.7& 67.2$\pm$1.9 & 67.4$\pm$3.0 \\
		\hline
		DaNN~\cite{ghifary2014domain}  & 35.0$\pm$0.2& 70.5$\pm$0.0 & 74.3$\pm$0.0 \\
		\hline
		Dlid~\cite{chopra2013dlid}  &26.1& 68.9 & 84.9 \\
		\hline
		Decaf~\cite{donahue2013decaf}  & 52.20$\pm$1.70& 91.50$\pm$1.50 & $-$ \\
		\hline
		DDC~\cite{tzeng2014deep}  & 59.40 $\pm$ 0.80 & 92.50$\pm$0.30 & 91.70$\pm$0.80 \\
		\hline
		\hline
	FCR2-c & 54.21$\pm$1.59 &	91.55$\pm$0.64 &	90.44$\pm$0.66 \\
	\hline
	FCR1-c  & 53.46$\pm$0.94 &	91.35$\pm$0.61 &	91.24$\pm$1.25 \\
	\hline 	 	 
	MLR  & 40.43$\pm$3.03 &	89.79$\pm$1.32 &	89.68$\pm$1.21 \\ 
	\hline 	 	 
	CFV & 21.06$\pm$2.72 &	76.50$\pm$1.32 &	80.08$\pm$1.13 \\
	\hline 	 	 
	FCR1-c+FCR2-c & $\textbf{55.17}$$\pm$$\textbf{1.91}$ &	92.53$\pm$0.78 &	92.21$\pm$1.06 \\
	\hline  	 	 
	MLR+FCR2-c &  52.28$\pm$2.61 &	93.36$\pm$1.58 &	93.49$\pm$0.96 \\
	\hline 	 	 
	CFV+FCR2-c & 49.36$\pm$1.92 &	91.80$\pm$0.91 &	92.77$\pm$0.97 \\
	\hline 	 	 
	MLR+FCR1-c & 49.46$\pm$3.03 &	92.53$\pm$1.41 &	92.89$\pm$1.17 \\
	\hline 	 	 
	CFV+FCR1-c	& 45.74$\pm$1.51 &	91.27$\pm$0.78 &	92.05$\pm$1.18 \\
	\hline  	 	 
	MLR+CFV & 36.15$\pm$3.72 &	90.79$\pm$1.00 &	90.20$\pm$1.22 \\
	\hline 	 	 
	MLR+CFV+FCR1-c & 45.21$\pm$3.43 &	92.98$\pm$1.19 &	93.29$\pm$1.45 \\
	\hline  	 	 
	MLR+CFV+FCR2-c & 49.33$\pm$3.53 &	93.18$\pm$1.16 &	93.57$\pm$1.40 \\
	\hline  	 	  
	CFV+FCR1-c+FCR2-c & 52.25$\pm$1.98 &	92.83$\pm$1.02 &	93.41$\pm$1.27 \\
	\hline  	 	 
	MLR+FCR1-c+FCR2-c & $\textbf{54.29}$$\pm$$\textbf{2.96}$ &	$\textbf{93.66}$$\pm$$\textbf{1.51}$ &	93.98$\pm$0.95 \\
	\hline   
	MLR+CFV+FCR1-c+FCR2-c & 51.45$\pm$3.33 &	93.46$\pm$1.26 &	$\textbf{94.22}$$\pm$$\textbf{1.36}$ \\
	\hline
	\end{tabular}}
	\vspace{-0.5cm}
	\label{table4}
\end{table}

\begin{table}[!t]
	\renewcommand{\arraystretch}{1}
	\caption{Classification accuracy of 31-category Office dataset under semi-supervised adaptation settings. 20 samples per class are used with Amazon as source domain, and 8 samples per class with Webcam and Dslr as the source domain respectively.}
	\label{table_example}
	\centering
    \scriptsize{\begin{tabular}{|c|c|c|c|}
		\hline
		\textbf{Methods}& $A\rightarrow W$ & $D\rightarrow W$ & $W\rightarrow D$\\
		\hline
		\hline
		GFK~(PLS,PCA)~\cite{gong2012geodesic}  & 46.4$\pm$0.5& 61.3$\pm$0.4 & 66.3$\pm$0.4 \\
		\hline
		SA~\cite{fernando2013unsupervised}  & 45.0& 64.8 & 69.9 \\
		\hline
		DaNBNN~\cite{tommasi2013frustratingly}  & 52.8$\pm$3.7& 76.6$\pm$1.7 & 76.2$\pm$2.5 \\
		\hline
		DaNN~\cite{ghifary2014domain}  & 53.6$\pm$0.2& 71.2$\pm$0.0 & 83.5$\pm$0.0 \\
		\hline
		Dlid~\cite{chopra2013dlid}  & 51.9& 78.2 & 89.9 \\
		\hline
		Decaf S+T~\cite{donahue2013decaf}  & 80.7$\pm$2.3& 94.8$\pm$1.2 & $-$ \\
		\hline
		DDC~\cite{tzeng2014deep}  & 84.1 $\pm$ 0.6 & 95.4$\pm$0.4 & 96.3$\pm$0.3 \\
		\hline
		DAN~\cite{Menglong2015deep}  & 85.6 $\pm$ 0.3 & 95.8$\pm$0.2 & 96.7$\pm$0.2 \\
		\hline
		\hline
			FCR2-c & 80.68$\pm$1.50 &	94.81$\pm$0.93 &	94.02$\pm$1.41 \\
			\hline
			FCR1-c  & 80.06$\pm$1.23 &	95.01$\pm$0.57 &	95.21$\pm$0.67 \\
			\hline 	 	 
			MLR  & 79.32$\pm$1.09 &	94.25$\pm$0.80 &	94.27$\pm$1.15 \\
			\hline 	 	 
			CFV & 73.05$\pm$1.86 &	87.52$\pm$1.42 &	88.59$\pm$1.88 \\
			\hline 	 	 
			FCR1-c+FCR2-c & 81.40$\pm$1.63 &	95.27$\pm$0.87 &	95.51$\pm$1.08 \\
			\hline  	 	 
			MLR+FCR2-c &  83.96$\pm$1.66 &	95.95$\pm$ 0.65 &	96.84$\pm$0.44 \\
			\hline 	 	 
			CFV+FCR2-c & 83.93$\pm$1.94 &	95.21$\pm$1.07 &	95.85$\pm$0.75 \\
			\hline 	 	 
			MLR+FCR1-c & 82.74$\pm$1.37 &	95.78$\pm$0.54 &	96.74$\pm$0.64 \\
			\hline 	 	 
			CFV+FCR1-c	& 82.56$\pm$1.79 &	94.62$\pm$0.97 &	96.05$\pm$0.82 \\
			\hline  	 	 
			MLR+CFV & 81.42$\pm$1.22 &	94.79$\pm$0.87 &	95.06$\pm$1.34 \\
			\hline 	 	 
			MLR+CFV+FCR1-c & 84.16$\pm$0.96 &	95.61$\pm$0.78 &	96.44$\pm$0.85 \\
			\hline  	 	 
			MLR+CFV+FCR2-c & $\textbf{84.79}$$\pm$$\textbf{1.62}$ &	95.98$\pm$0.50 &	96.89$\pm$0.37 \\
			\hline  	 	  
			CFV+FCR1-c+FCR2-c & 83.50$\pm$2.06 &	95.24$\pm$0.64 &	96.54$\pm$0.84 \\
			\hline  	 	 
			MLR+FCR1-c+FCR2-c & 83.59$\pm$1.45 &	$\textbf{96.15}$$\pm$$\textbf{0.48}$ &	96.94$\pm$0.64 \\
			\hline   
			MLR+CFV+FCR1-c+FCR2-c & 84.64$\pm$1.57 &	95.95$\pm$0.54 &	$\textbf{97.14}$$\pm$$\textbf{0.37}$ \\
			\hline
	\end{tabular}}
	\vspace{-0.5cm}
	\label{table5}
\end{table}

\begin{table}[!t]
	\renewcommand{\arraystretch}{1}
	\caption{Classification accuracy of 31-category Office dataset under unsupervised adaptation settings. All source domain data are used for training.}
	\label{table_example}
	\centering
	\scriptsize{\begin{tabular}{|c|c|c|c|}
		\hline
		\textbf{Methods}& $A\rightarrow W$ & $D\rightarrow W$ & $W\rightarrow D$\\
		\hline
		\hline
		GFK~~\cite{Menglong2015deep}  & 21.4$\pm$0.2& 69.1$\pm$0.3 & 65.0$\pm$0.2 \\
		\hline
        CNN~\cite{Menglong2015deep}  & 59.40$\pm$0.5& 94.40$\pm$0.3 & 98.80$\pm$0.2 \\
		\hline
		LapCNN~\cite{Menglong2015deep}  & 60.40 $\pm$ 0.30 & 94.70$\pm$0.50 & 99.10$\pm$0.20 \\
		\hline
		DDC~\cite{Menglong2015deep}  & 60.50 $\pm$ 0.70 & 94.80$\pm$0.50 & 98.50$\pm$0.40 \\
		\hline
		DAN~\cite{Menglong2015deep}  & 64.50 $\pm$ 0.40 & 95.20$\pm$0.30 & 98.60$\pm$0.20 \\
		\hline
		\hline
		FCR2-c & 61.38 & 94.72 & 98.80 \\
		\hline
		FCR1-c & 59.50 & 95.72 & 99.40 \\
		\hline
		MLR  & 50.31 &	93.08 & 97.79 \\
		\hline
		CFV & 25.91 & 82.52	& 92.17 \\
		\hline
		FCR1-c+FCR2-c & 63.02 & 95.85 & 99.60 \\
		\hline
		MLR+FCR2-c & 61.51	& 96.10 & 99.60 \\
		\hline
		CFV+FCR2-c & 55.22 & 96.35	& 99.40 \\
		\hline
		MLR+FCR1-c & 57.74 & 95.85 & 99.20 \\
		\hline
		CFV+FCR1-c	 & 51.57 &	95.47 &	99.60 \\
		\hline
		MLR+CFV & 44.78 & 94.72 & 97.99 \\
		\hline
		MLR+CFV+FCR1-c & 53.33 & 96.48 & 99.00 \\
		\hline
		MLR+CFV+FCR2-c & 56.60 & 96.48 & 99.40 \\
		\hline
		CFV+FCR1-c+FCR2-c &	58.87 &	$\textbf{96.73}$ & 99.60 \\
		\hline
		MLR+FCR1-c+FCR2-c &	$\textbf{63.52}$ &	$\textbf{96.73}$ & $\textbf{99.80}$ \\
		\hline
		MLR+CFV+FCR1-c+FCR2-c & 59.50 &	96.48 &	$\textbf{99.80}$ \\
		\hline
	\end{tabular}}
	\vspace{-0.3cm}
	\label{table6}
\end{table}

 \begin{table}[!t]
 	\renewcommand{\arraystretch}{1}
 	\caption{Classification accuracy of 31-category Office dataset under semi-unsupervised adaptation settings. All source domain data and three more target samples per class are added for training.}
 	
 	\centering
 	\scriptsize{\begin{tabular}{|c|c|c|c|}
 		\hline
 		\textbf{Methods}& $A\rightarrow W$ & $D\rightarrow W$ & $W\rightarrow D$\\
 		\hline 
 		\hline  
 			FCR2-c & 76.27$\pm$0.59 & 96.18$\pm$0.48 & 98.86$\pm$0.37 \\
 			\hline
 			FCR1-c & 77.75$\pm$0.87 & 96.87$\pm$0.66 & 99.36$\pm$0.22 \\
 			\hline 	 	 
 			MLR  & 77.69$\pm$1.39 &	96.21$\pm$0.72 & 99.01$\pm$0.60 \\
 			\hline 	 	 
 			CFV & 69.86$\pm$1.58 & 91.99$\pm$1.38	& 94.72$\pm$1.15 \\
 			\hline 	 	 
 			FCR1-c+FCR2-c & 80.26$\pm$1.42 & 97.12$\pm$0.31 & 99.56$\pm$0.11 \\
 			\hline  	 	 
 			MLR+FCR2-c & 83.65$\pm$1.53	& 97.89$\pm$0.31 & 99.80$\pm$0.21 \\
 			\hline 	 	 
 			CFV+FCR2-c & 83.13$\pm$1.19 & 97.61$\pm$0.56	& 99.65$\pm$0.22 \\
 			\hline 	 	 
 			MLR+FCR1-c & 82.28$\pm$1.73 & 97.64$\pm$0.42 & 99.95$\pm$0.11 \\
 			\hline 	 	 
 			CFV+FCR1-c	 & 81.40$\pm$1.90 &	97.38$\pm$0.52 &	99.80$\pm$0.21 \\
 			\hline  	 	 
 			MLR+CFV & 80.34$\pm$1.95 & 97.04$\pm$0.86 & 98.86$\pm$0.81 \\
 			\hline 	 	 
 			MLR+CFV+FCR1-c & 83.56$\pm$1.82 & 97.81$\pm$0.57 & 99.95$\pm$0.11 \\
 			\hline  	 	 
 			MLR+CFV+FCR2-c & $\textbf{84.30}$$\pm$$\textbf{1.74}$ & $\textbf{98.12}$$\pm$$\textbf{0.53}$ & 99.90$\pm$0.22 \\
 			\hline  	 	  
 			CFV+FCR1-c+FCR2-c &	83.33$\pm$1.78 & 97.75$\pm$0.23 & 99.95$\pm$0.11 \\
 			\hline  	 	 
 			MLR+FCR1-c+FCR2-c &	83.65$\pm$1.63 &	98.06$\pm$0.31 & $\textbf{100.0}$$\pm$$\textbf{0.00}$ \\
 			\hline   
 			MLR+CFV+FCR1-c+FCR2-c & $\textbf{85.24}$$\pm$$\textbf{1.98}$ &	98.09$\pm$0.42 &	$\textbf{100.0}$$\pm$$\textbf{0.00}$ \\
 			\hline
 	\end{tabular}}
 	\label{table7}
 	\vspace{-0.5cm}
 \end{table}
\subsection{Parameter Analysis}
{\color{black}In this part, we take Indoor-67 as an example to evaluate the influence of the key parameters $(\lambda_B,\lambda_F)$~(in Eq.~\ref{Eq3}) w.r.t. the final accuracy of the generated MLR, under AlexNet. Specifically, we evaluate eight combinations which are listed in the first column of Table~\ref{table8}. The results of MLR in Table~\ref{table8} are obtained under the class-mixture dictionary with size of 10$\times$$\textbf{Category number}$, and CFV and FCRs are obtained under AlexNet, the same as the original experimental settings. It can be seen from Table~\ref{table8} that the best accuracy of MLR is achieved under $(\lambda_B,\lambda_F)=(1,0)$, while the best accuracy~($72.94\%$) of hybrid representation~(MLR+CFV+FCR1-w) is achieved under $(\lambda_B,\lambda_F)=(0.7,0.3)$, which is only slightly better than the result~($72.89\%$) under $(\lambda_B,\lambda_F)=(1,0)$. It can be concluded that (1) MLR is slightly sensitive to the values of $(\lambda_B,\lambda_F)$, the best result is obtained under $\lambda_F=0$ which further validates the conclusion that feature space information is not necessary during clustering proposals of each image for dense layout database of Indoor-67. (2) The final best hybrid representation~(MLR+CFV+FCR1-w) is actually robust to the values of $(\lambda_B,\lambda_F)$, which is very important since we cannot try too many combinations of $(\lambda_B,\lambda_F)$ in the real applications. Therefore, the hybrid representation of MLR+CFV+FCRs~(therein, we set $(\lambda_B,\lambda_F)=(1,0)$, which makes MLR have the best performance) should be still an effective final features to train classifiers and conduct predictions.
}

\begin{table*}[!t]
	\renewcommand{\arraystretch}{1}
	\caption{The classification rate of MLR and its combinations with other representations with class-mixture dictionary size of $10\times \textbf{(number of categories)}$, under AlexNet on Indoor-67.}
	\label{table_example}
	\centering
	\scriptsize{\begin{tabular}{|c|c|c|c|c|c|c|c|c|c|c|c|}
			\hline
			{Methods} / ($\lambda_B,\lambda_F$)& (1,0) &	(0.9,0.1) &	(0.8,0.2) & (0.7,0.3) &	(0.6,0.4) &	(0.5,0.5) &	(0.4,0.6) &	(0.3,0.7) &	(0.2,0.8) &	(0.1,0.9) &	(0,1)\\
			\hline 
			
			MLR  & \bf {\color{black}67.72}&	66.72&	66.54&	66.65&	65.85&	65.64&	66.67&	65.27&	66.18&	66.66&	66.55 \\
			\hline 	 	 
			MLR+FCR2-w & 68.58&	68.45&	68.38&	68.43&	67.61&	68.53&	68.20&	68.07&	67.07&	68.42&	68.41  \\
			\hline 	 	 	 	 
			MLR+FCR1-w & 68.71&	68.87&	68.93&	68.56&	68.70&	68.79&	68.78&	67.99&	68.13&	68.26&	68.84 \\
			\hline 	 	 
			MLR+CFV & 71.43&	71.13&	71.09&	71.01&	71.18&	71.25&	70.77&	71.23&	70.62&	70.41&	70.46 \\
			\hline 	 	 
			MLR+CFV+FCR1-w & \bf {\color{black}72.89}&	72.56&	72.48& \bf {\color{black}72.94}&	71.81&	\bf {\color{black}72.77}&	72.10&	\bf {\color{black} 72.91}&	72.24&	71.87&	72.04 \\
			\hline  	 	 
			MLR+CFV+FCR2-w & 72.67&	72.25&	71.80&	72.21&	71.52&	71.96&	71.36&	71.51&	71.90&	72.46&	71.86  \\
			\hline
			MLR+FCR1-w+FCR2-w &	68.98&	68.62&	68.17&	67.70&	68.09&	68.90&	68.14&	67.77&	68.18&	68.25&	68.47\\
			\hline   
			MLR+CFV+FCR1-w+FCR2-w & 72.01&	71.91&	71.64&	71.72&	71.38&	71.91&	71.45&	71.17&	71.21&	71.67&	71.23 \\
			\hline
		\end{tabular}}
		\vspace{-0.4cm}
		\label{table8}
	\end{table*}
\subsection{Complementarity with Other Nets}

{\color{black}In this part, we combine our proposed hybrid representation ~(from VGG-19) with the representations~(from GoogLeNet~\cite{szegedy2014going} and/or VGG-11~\cite{WangGHQ15} trained on Place205 database\cite{zhou2014learning}) to verify their complementarity to each other. For GoogLeNet, given the input resized image of size $224\times224$, the last convolutional layer is first extracted and followed by average pooling to output a 1024 dimensional vector, which is further $L$2 normalized. For VGG-11, the first fully connected layer representations are extracted and $L$2 normalized. The results on both Indoor-67 and SUN-397 database are listed in Table~\ref{Table9}. It can be concluded that our proposed hybrid representation can greatly enhance the performance when combined with GoogLeNet and/or VGG-11 representations. As far as we know, the performances of 85.97\% and 70.69\% are the best results on Indoor-67 and SUN-397 databases respectively, which are much better than current state-of-the-art performance.
}
 \begin{table}[!t]
 	\renewcommand{\arraystretch}{1}
 	\caption{Complementary classification Rate of VGG-19, VGG-11 and GoogLeNet.}
 	\centering
 	\scriptsize{\begin{tabular}{|c|c|c|c|}
 		\hline
 		Methods & Indoor-67~(\%) &  SUN-397~(\%)\\
 		\hline
 		G\_P205 & 76.84  & 63.39$\pm$0.22   \\
 		\hline
 		VGG-11\_P205 &  82.74 & 66.71$\pm$0.08   \\
 		\hline
 		VGG-19\_Hybrid & 82.24 & 64.53$\pm$0.24 \\
 		\hline
 		VGG-11\_P205+G\_P205 & 83.42  & 68.01$\pm$0.24   \\
 		\hline
 		VGG-19\_Hybrid+VGG-11\_P205 & 85.59  & 69.55$\pm$0.13   \\
 		\hline
 		VGG-19\_Hybrid+G\_P205 & 84.91 & 69.48$\pm$0.17  \\
 		\hline
 	    VGG-19\_Hybrid+VGG-11\_P205+G\_P205 & $\textbf{85.97}$ & \bf 70.69$\pm$0.15  \\
 	    \hline
 	\end{tabular}}
 	\vspace{-0.5cm}
 	\label{Table9}
 \end{table} 
\subsection{Time Complexity Analysis}
As our representations consist of three parts, i.e., MLR, CFV and FCR, the total time consumptions will be the sum of constructing MLR, CFV and FCR. Specifically, time consumption of MLR is mainly due to the two stage clustering~($T_{1}$) and and approximated LLC coding~\cite{wang2010locality} for generating different feature maps~($T_{2}$). And the time consumption of CFV is the GMM training~($T_{3}$) and Fisher vector coding~($T_{4}$). Finally, time consumption of FCR is the time of forward propagation of CNN~($T_{5}$). So the total time consumption for constructing our representation is $T=\sum_{i=1}^{5}T_{i}$. After obtaining the representations of all the training and testing images, we train linear SVM classifier. The time consumption for training linear SVM and testing should also be considered for the total time complexity. In our experiments, $T$ is approximately two hours for SUN-397 database and less than one hour for Indoor-67 and Office databases.      

\section{Conclusion And Future Works}
In this paper, we propose a hybrid representation method for scene recognition and domain adaptation by integrating the powerful CNN features with the traditional well-studied dictionary-based features. An efficient two-stage part dictionary generating method is used to exploit local discriminative and structural information. Based on the generated class-specific and class-mixture part dictionaries, we can get a novel mid-level local representation~(denoted as MLR) containing rich local discriminative information, which is not considered during the CNN training process. Moreover, the Fisher vector representation of convolutional layer~(CFV) is further used to boost the accuracy. The global representation of the fully connected layer~(denoted as FCR) of CNN is validated to be a powerful representation. By combining the complementary information in MLR, CFV, and FCR, a hybrid feature representation can be generated which is much more accurate than the traditional CNN features. Experiments on scene recognition and domain adaptation demonstrate the excellent performance of our hybrid models. In future, to further boost the accuracy, we will jointly train the CNN model and MLR~(CFV) under a unified framework. {\color{black} Another interesting future work is to construct our hybrid representations based on the networks trained on place database.}

\section*{Acknowledgment}

This work was supported by the National Basic Research Program of China (973 Program) Grant 2012CB316302, the Strategic Priority Research Program of the CAS (Grant XDA06040102) and National Natural Science Foundation of China (NSFC) Grant 61403380.

\ifCLASSOPTIONcaptionsoff
  \newpage
\fi

\bibliographystyle{IEEEtran}
\bibliography{mybibfile}
\vspace{-1.4cm}
\begin{IEEEbiography}[{\includegraphics[width=1in,height=1.4in,clip,keepaspectratio]{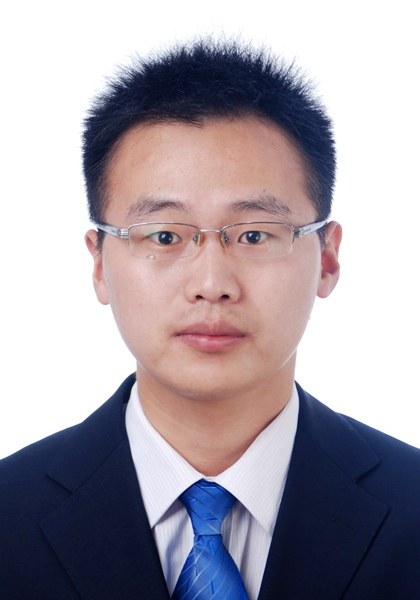}}]{Guo-Sen Xie}
is currently pursuing his Ph.D. degree in pattern recognition and intelligent systems, at the National Laboratory of Pattern Recognition, Institute of Automation, Chinese Academy of Sciences, Beijing, China. His research interests include machine learning, deep learning, and their applications to object recognition and DNA sequence analysis.
\end{IEEEbiography}
\vspace{-1.3cm}
\begin{IEEEbiography}[{\includegraphics[width=1in,height=1.5in,clip,keepaspectratio]{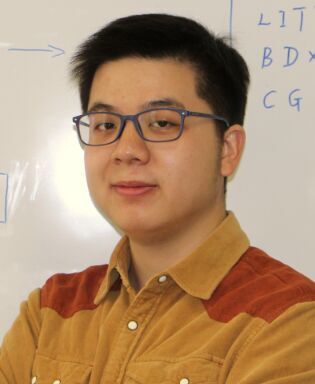}}]{Xu-Yao Zhang}
received the BS degree in computational mathematics from Wuhan University, Wuhan, China, in 2008, and the PhD degree in pattern recognition and intelligent systems from Institute of Automation, Chinese Academy of Sciences, Beijing, China, in 2013. From July 2013, he has been an Assistant Professor at the National Laboratory of Pattern Recognition, Institute of Automation, Chinese Academy of Sciences, Beijing, China. His research interests include machine learning, pattern recognition, handwriting recognition, and deep learning.
\end{IEEEbiography}
\vspace{-1.6cm}
\begin{IEEEbiography}[{\includegraphics[width=0.88in,height=1.1in]{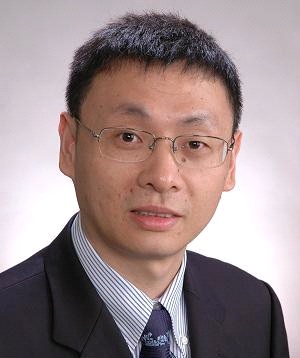}}]{Shuicheng Yan} is currently an Associate Professor at the Department of Electrical and Computer Engineering at National University of Singapore, and the founding lead of the Learning and Vision Research Group (http://www.lv-nus.org). Dr. Yan's research areas include machine learning, computer vision and multimedia, and he has authored/co-authored hundreds of technical papers over a wide range of research topics, with Google Scholar citation $>$ 19,000 times and H-index 51. He is ISI Highly-cited Researcher, 2014 and IAPR Fellow 2014. He has been serving as an associate editor of IEEE TKDE, TCSVT and ACM Transactions on Intelligent Systems and Technology (ACM TIST). He received the Best Paper Awards from ACM MM'13 (Best Paper and Best Student Paper), ACM MM’12 (Best Demo), PCM'11, ACM MM’10, ICME’10 and ICIMCS'09, the runner-up prize of ILSVRC'13, the winner prize of ILSVRC’14 detection task, the winner prizes of the classification task in PASCAL VOC 2010-2012, the winner prize of the segmentation task in PASCAL VOC 2012, the honourable mention prize of the detection task in PASCAL VOC'10, 2010 TCSVT Best Associate Editor (BAE) Award, 2010 Young Faculty Research Award, 2011 Singapore Young Scientist Award, and 2012 NUS Young Researcher Award.
\end{IEEEbiography}
\vspace{-1.6cm}
\begin{IEEEbiography}[{\includegraphics[width=1in,height=1.5in,clip,keepaspectratio]{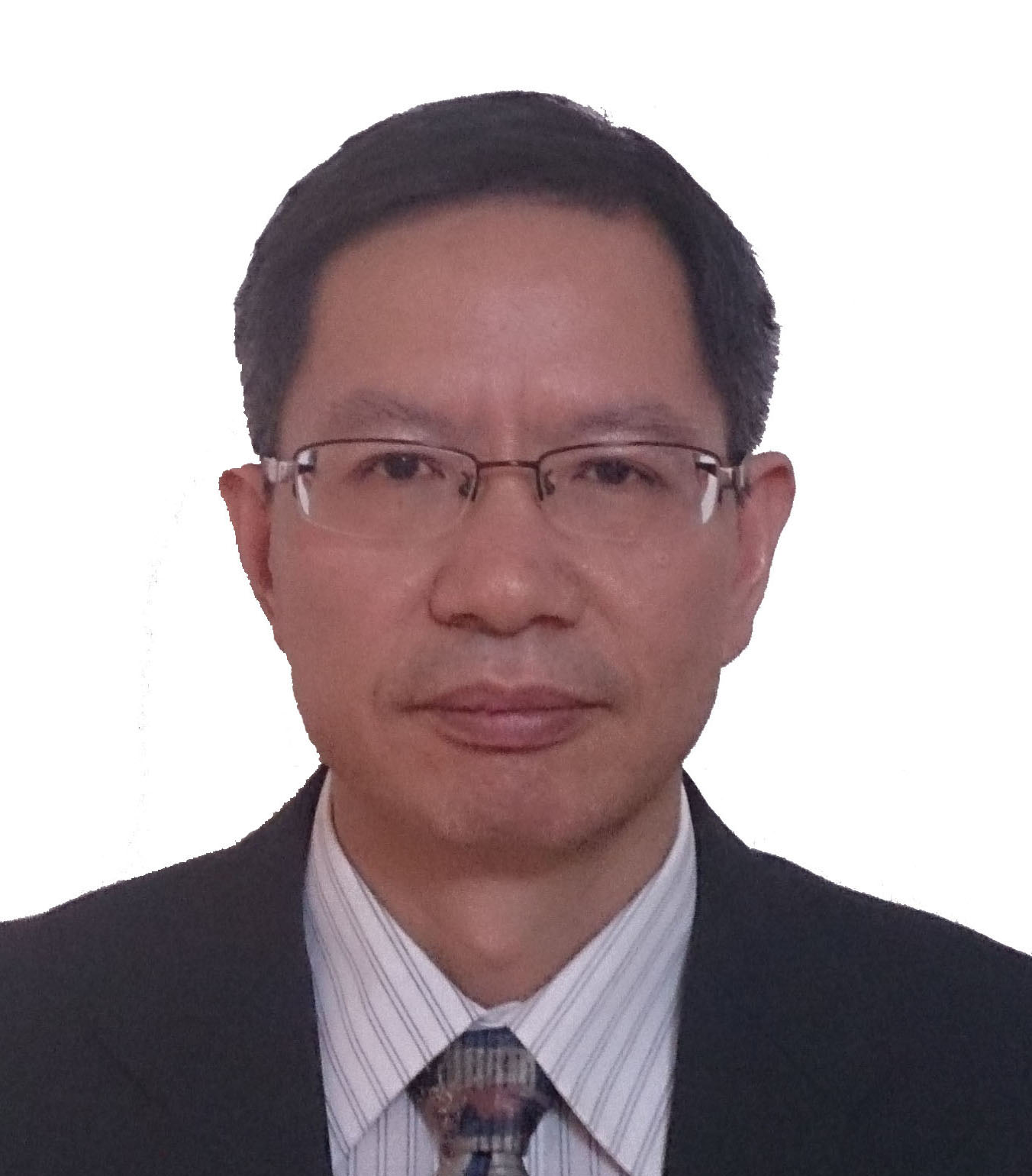}}]{Cheng-Lin Liu}
received the B.S. degree in Electronic Engineering from Wuhan University, Wuhan, China, the M.E. degree in Electronic Engineering from Beijing Polytechnic University (current Beijing University of Technology), Beijing, China, the Ph.D. degree in Pattern Recognition and Intelligent Systems from the Institute of Automation of Chinese Academy of Sciences, Beijing, China, in 1989, 1992 and 1995, respectively. He was a postdoctoral fellow at Korea Advanced Institute of Science and Technology (KAIST) and later at Tokyo University of Agriculture and Technology from March 1996 to March 1999. From 1999 to 2004, he was a research staff member and later a senior researcher at the Central Research Laboratory, Hitachi, Ltd., Tokyo, Japan. From 2005, he has been a Professor at the National Laboratory of Pattern Recognition (NLPR), Institute of Automation of Chinese Academy of Sciences, Beijing, China, and is now the Director of the laboratory. His research interests include pattern recognition, image processing, neural networks, machine learning, and the applications to character recognition and document analysis. He has published over 200 technical papers at prestigious international journals and conferences. He is a Fellow of the IAPR and the IEEE.
\end{IEEEbiography}

\end{document}